# Learning to Make Predictions In Partially Observable Environments Without a Generative Model

**Erik Talvitie**                                    ERIK.TALVITIE@FANDM.EDU
*Mathematics and Computer Science*
*Franklin and Marshall College*
*Lancaster, PA 17604-3003, USA*

**Satinder Singh**                                    BAVEJA@UMICH.EDU
*Computer Science and Engineering*
*University of Michigan*
*Ann Arbor, MI 48109-2121, USA*

## Abstract

When faced with the problem of learning a model of a high-dimensional environment, a common approach is to limit the model to make only a restricted set of predictions, thereby simplifying the learning problem. These partial models may be directly useful for making decisions or may be combined together to form a more complete, structured model. However, in partially observable (non-Markov) environments, standard model-learning methods learn generative models, i.e. models that provide a probability distribution over all possible futures (such as POMDPs). It is not straightforward to restrict such models to make only certain predictions, and doing so does not always simplify the learning problem. In this paper we present prediction profile models: non-generative partial models for partially observable systems that make only a given set of predictions, and are therefore far simpler than generative models in some cases. We formalize the problem of learning a prediction profile model as a transformation of the original model-learning problem, and show empirically that one can learn prediction profile models that make a small set of important predictions even in systems that are too complex for standard generative models.

## 1. Introduction

Learning a model of the dynamics of an environment through experience is a critical capability for an artificial agent. Agents that can learn to make predictions about future events and anticipate the consequences of their own actions can use these predictions to plan and make better decisions. When the agent's environment is very complex, however, this learning problem can pose serious challenges. One common approach to dealing with complex environments is to learn *partial models*, focusing the model-learning problem on making a restricted set of particularly important predictions. Often when only a few predictions need to be made, much of the complexity of the dynamics being modeled can be safely ignored. Sometimes a partial model can be directly useful for making decisions, for instance if the model makes predictions about the agent's future rewards (e.g., see McCallum, 1995; Mahmud, 2010). In other cases, many partial models making restricted predictions are combined to form a more complete model as in, for instance, factored MDPs (Boutilier, Dean, & Hanks, 1999), factored PSRs (Wolfe, James, & Singh, 2008), or "collections of local models" (Talvitie & Singh, 2009b).

The most common approach to learning a partial model is to apply an *abstraction* (whether learned or supplied by a domain expert) that filters out detail from the training data that is irrelevant to making the important predictions. Model-learning methods can then be applied to the abstract data, and typically the learning problem will be more tractable as a result. However, especially in the case of partially observable systems, abstraction alone may not sufficiently simplify the learning problem, even (as we will see in subsequent examples) when the model is being asked to make intuitively simple predictions. The counter-intuitive complexity of learning a partial model in the partially observable case is a direct result of the fact that standard model-learning approaches for partially observable systems learn *generative models* that attempt to make every possible prediction about the future and cannot be straightforwardly restricted to making only a few particularly important predictions.

In this paper we present an alternative approach that learns *non-generative* models that make *only* the specified predictions, conditioned on history. In the following illustrative example, we will see that sometimes a small set of predictions is all that is necessary for good control performance but that learning to make these predictions in a high-dimensional environment using standard generative models can pose serious challenges. By contrast we will see that there exists a simple, non-generative model that can make and maintain these predictions and this will form the learning target of our method.

## 1.1 An Example

Consider the simple game of Three Card Monte. The dealer, perhaps on a crowded street, has three cards, one of which is an ace. The dealer shows the location of the ace, flips over the cards, and then mixes them up by swapping two cards at every time step. A player of the game must keep track of location of the ace. Eventually the dealer stops mixing up the cards and asks for a guess. If a player correctly guesses where the ace is, they win some money. If they guess wrong, they lose some money.

Consider an artificial agent attempting to learn a model of the dynamics of this game from experience. It takes a sequence of actions and perceives a sequence of observations. The raw data received by the agent includes a rich, high-dimensional scene including the activities of the crowd, the movement of cars, the weather, as well as the game itself (the dealer swapping cards). Clearly, learning a model that encompasses all of these complex phenomena is both infeasible and unnecessary. In order to win the game, the agent needs only focus on making predictions about the cards, and need not anticipate the future behavior of the city scene around it. In particular, the agent need only make three predictions: "If I flip over card 1, will it be the ace?" and the corresponding predictions for cards 2 and 3. One can safely ignore much of the detail in the agent's experience and still make these important predictions accurately. Once one filters out the irrelevant detail, the agent's experience might look like this:

$$bet\ pos2\ watch\ swap1, 2\ watch\ swap2, 3\ \ldots,$$

where the agent takes the *bet* action, starting the game, and observes the dealer showing the card in position 2. Then the agent takes the *watch* action, observes the dealer swapping cards 1 and 2, takes the *watch* action again, observes the dealer swapping cards 2 and 3, and so on until the dealer prompts the agent for a guess (note that this is not an uncontrolled system; *watch* is indeed an action that the agent must select over, say, reaching out and flipping the cards itself, which in a real game of Three Card Monte would certainly result in negative utility!) Now the data reflects *only* the movement of the cards. One could learn a model using this new data set and the learning problem would be far simpler than before since complex and irrelevant phenomena like the crowd and the weather have been ignored.

In the Markov case, the agent directly observes the entire state of the environment and can therefore learn to make predictions as a direct function of state. Abstraction simplifies the representation of state and thereby simplifies the learning problem. Note, however, that the Three Card Monte problem is *partially observable* (non-Markov). The agent cannot directly observe the state of the environment (the location of the ace and the state of the dealer's mind are both hidden to the agent). In the partially observable case, the agent must *learn* to maintain a compact representation of state as well as learn the dynamics of that state. The most common methods to achieve this, such as expectation-maximization (EM) for learning POMDPs (Baum, Petrie, Soules, & Weiss, 1970), learn *generative models* which provide a probability distribution over all possible futures.

In Three Card Monte, even when all irrelevant details have been ignored and the data contains only information about the cards' movement, a generative model will *still* be intractably complex! A generative model makes predictions about *all* future events. This includes the predictions the model is meant to make (such as whether flipping over card 1 in the next time-step will reveal the ace) but also many irrelevant predictions. A generative model, will also predict, for instance, whether flipping over card 1 in 10 time-steps will reveal the ace or whether cards 1 and 2 will be swapped in the next time-step. To make these predictions, the model must capture not only the dynamics of the cards but also of the *dealer's decision-making process*. If the dealer decides which cards to swap using some complex process (as a human dealer might) then the problem of learning a generative model of this abstract system will be correspondingly complex.

Of course, in Three Card Monte, predicting the dealer's future behavior is entirely unnecessary to win. All that is required is to maintain the ace's *current* location over time. As such, learning a model that devotes most of its complexity to anticipating the dealer's decisions is counter-intuitive at best. A far more reasonable model can be seen in Figure 1. Here the "states" of the model are labeled with predictions about the ace's location. The transitions are labeled with observations of the dealer's behavior. As an agent plays the game, it could use such a model to maintain its predictions about the location of the ace over time, taking the dealer's behavior into account, but not *predicting* the dealer's future behavior. Note that this is a *non-generative* model. It does not provide a distribution over all possible futures and it cannot be used to "simulate the world" because it does not predict the dealer's next move. It only provides a limited set of conditional predictions about the future, given the history of past actions and observations. On the other hand, it is far simpler than a generative model would be. Because it does not model the dealer's decision-making process, this model has only 3 states, *regardless* of the underlying process used by the dealer.

The model in Figure 1 is an example of what we term a *prediction profile model*. This paper will formalize prediction profile models and present an algorithm for learning them from data, under some assumptions (to be specified once we have established some necessary terminology). We will empirically demonstrate that in some partially observable systems that prove too complex for standard generative model-learning methods, it is possible to learn a prediction profile model that makes a small set of important predictions that allow the agent to make good decisions. The next sections will formally describe the setting and establish some notation and terminology and formalize the general learning problem being addressed. Subsequent sections will formally present prediction profile models and an algorithm for learning them, as well as several relevant theoretical and empirical results.

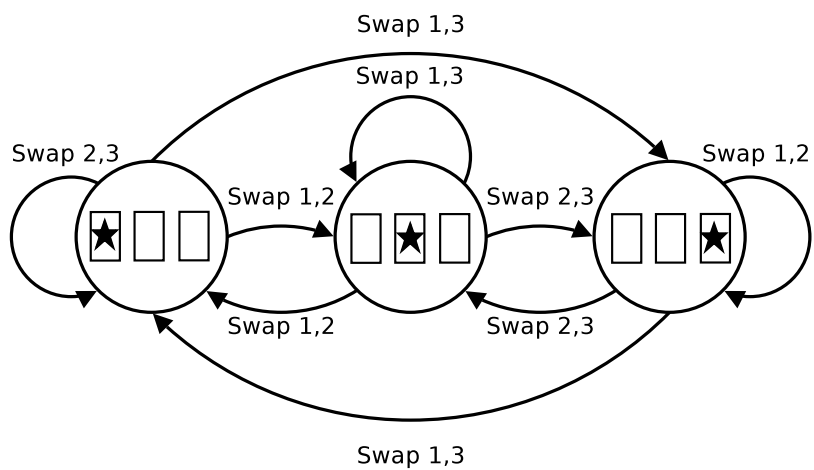

Figure 1: Maintaining predictions about the location of the ace in Three Card Monte. Transitions are labeled with the dealer's swaps. States are labeled with the predicted position of the special card.

## 1.2 Discrete Dynamical Systems

We focus on discrete dynamical systems. The agent has a finite set $\mathcal{A}$ of actions that it can take and the environment has a finite set $\mathcal{O}$ of observations that it can produce. At every time step $i$, the agent chooses an action $a^i \in \mathcal{A}$ and the environment stochastically emits an observation $o^i \in \mathcal{O}$.

**Definition 1.** At time step $i$, the sequence of past actions and observations since the beginning of time $h^i = a^1 o^1 a^2 o^2 \ldots a^i o^i$ is called the *history* at time $i$.

The history at time zero, before the agent has taken any actions or seen any observations $h^0$, is called the *null history*.

### 1.2.1 PREDICTIONS

An agent uses its model to make conditional predictions about future events, given the history of actions and observations *and* given its own future behavior. Because the environment is assumed to be stochastic, predictions are probabilities of future events. The primitive building block used to describe future events is called a *test* (after Rivest & Schapire, 1994; Littman, Sutton, & Singh, 2002). A test $t$ is simply a sequence of actions and observations that could possibly occur, $t = a_1 o_1 \ldots a_k o_k$. If the agent actually takes the action sequence in $t$ and observes the observation sequence in $t$, we say that test $t$ *succeeded*. A *prediction* $p(t \mid h)$ is the probability that test $t$ succeeds after history $h$, assuming the agent takes the actions in the test. Essentially, the prediction of a test is the answer to the question "If I were to take this particular sequence of actions, with what probability would I see this particular sequence of observations, given the history so far?" Formally,

$$p(t \mid h) \stackrel{\text{def}}{=} \Pr(o_1 \mid h, a_1)\Pr(o_2 \mid ha_1 o_1, a_2) \ldots \Pr(o_k \mid ha_1 o_1 a_2 o_2 \ldots a_{k-1} o_{k-1}, a_k). \quad (1)$$

Let $\mathcal{T}$ be the set of all tests (that is, the set of all possible action-observation sequences of all lengths). Then the set of all possible histories $\mathcal{H}$ is the set of all action-observation sequences that could possibly occur starting from the null history, and the null history itself: $\mathcal{H} \stackrel{\text{def}}{=} \{t \in \mathcal{T} \mid p(t \mid h^0) > 0\} \cup \{h^0\}$.

A model that can make a prediction $p(t \mid h)$ for *all* $t \in \mathcal{T}$ and $h \in \mathcal{H}$ can make *any* conditional prediction about the future (Littman et al., 2002). Because it represents a probability distribution over all futures, such a model can be used to sample from that distribution in order to "simulate the world," or sample possible future trajectories. As such, we call a model that makes all predictions a *generative model*.

Note that the use of the word "generative" here is closely related to its broader sense in general density estimation. If one is attempting to represent the conditional probability distribution $\Pr(A \mid B)$, the *generative* approach would be to represent the full joint distribution $\Pr(A, B)$ from which the conditional probabilities can be computed as $\frac{\Pr(A,B)}{\Pr(B)}$. That is to say, a generative model in this sense makes predictions even about variables we only wish to condition on. The *non-generative* or, in some settings, *discriminitive* approach would instead directly represent the conditional distribution, taking the value of $B$ as un-modeled input. The non-generative approach can sometimes result in significant savings if $\Pr(B)$ is very difficult to represent/learn, but $\Pr(A \mid B)$ is relatively simple (so long as one is truly disinterested in modeling the joint distribution).

In our particular setting, a generative model is one that provides a probability distribution over all futures (given the agent's actions). As such, one would use a generative model to compute $p(t \mid h)$ for some particular $t$ and $h$ as $\frac{p(ht \mid h^0)}{p(h \mid h^0)}$. In fact, from Equation 1 one can see that the prediction for any multi-step test can be computed from the predictions of one-step tests:

$$p(a_1 o_1 a_2 o_2 \ldots a_k o_k \mid h) = p(a_1 o_1 \mid h)p(a_2 o_2 \mid ha_1 o_1) \ldots p(a_k o_k \mid ha_1 o_1 a_2 o_2 \ldots a_k o_k).$$

This leads to a simple definition of a generative model:

**Definition 2.** Any model that can provide the predictions $p(ao \mid h)$ for all actions $a \in \mathcal{A}$, observations $o \in \mathcal{O}$ and histories $h \in \mathcal{H}$ is a *generative model*.

A non-generative model, then, would not make all one-step predictions in all histories and, consequently, would have to directly represent the prediction $p(t \mid h)$ with the history $h$ as an un-modeled input. It would condition on a given history, but not necessarily be capable of computing the probability of that history sequence. As we saw in the Three Card Monte example, this can be beneficial if making and maintaining predictions for $t$ is substantially simpler than making predictions for every possible action-observation sequence.

Note that a test describes a very specific future event (a sequence of specific actions and observations). In many cases one might wish to make predictions about more abstract events. This can be achieved by composing the predictions of many tests. For instance *set tests* (Wingate, Soni, Wolfe, & Singh, 2007) are a sequence of actions and a *set* of observation sequences. A set test succeeds when the agent takes the specified action sequence and sees *any* observation sequence contained within the set occur. While traditional tests allow an agent, for instance to express the question "If I go outside, what is the probability I will see this exact sequence of images?" a set test can express the far more useful, abstract question "If I go outside, what is the probability that it will be sunny?" by grouping together all observations of a sunny day. Even more generally, *option tests* (Wolfe & Singh, 2006; Soni & Singh, 2007) express future events where the agent's behavior is described abstractly as well as the resulting observations. These types of abstract predictions can be computed as the linear combination of a set of concrete predictions.

### 1.2.2 SYSTEM DYNAMICS MATRIX AND LINEAR DIMENSION

It is sometimes useful to describe a dynamical system using a conceptual object called the *system dynamics matrix* (Singh, James, & Rudary, 2004). The system dynamics matrix contains the values of *all possible* predictions, and therefore fully encodes the dynamics of the system. Specifically,

**Definition 3.** The *system dynamics matrix* of a dynamical system is an infinity-by-infinity matrix. There is a column corresponding to every test $t \in \mathcal{T}$. There is a row corresponding to every history $h \in \mathcal{H}$. The $ij$th entry of the system dynamics matrix is the prediction $p(t_j \mid h_i)$ of the test corresponding to column $j$ at the history corresponding to row $i$ and there is an entry for every history-test pair.

Though the system dynamics matrix has infinitely many entries, in many cases it has finite rank. The rank of the system dynamics matrix can be thought of as a measure of the complexity of the system (Singh et al., 2004).

**Definition 4.** The *linear dimension* of a dynamical system is the rank of the corresponding system dynamics matrix.

For some popular modeling representations, the linear dimension is a major factor in the complexity of representing and learning a generative model of the system. For instance, in POMDPs, the number of hidden states required to represent the system is lower-bounded by the linear dimension. In this work we adopt linear dimension as our measure of the complexity of a dynamical system. When we say a system is "simpler" than another, we mean it has a lower linear dimension.

### 1.2.3 THE MARKOV PROPERTY

A dynamical system is *Markov* if all that one needs to know about history in order to make predictions about future events is the most recent observation.

**Definition 5.** A system is *Markov* if for any two histories $h$ and $h'$ (that may be the null history), any two actions $a$ and $a'$, any observation $o$, and any test $t$, $p(t \mid hao) = p(t \mid h'a'o)$.

In the Markov case we will use the notational shorthand $p(t \mid o)$ to indicate the prediction of $t$ at *any* history that ends in observation $o$. In the Markov case, because observations contain all the information needed to make any prediction about the future, they are often called *state* (because they describe the state of the world). When a system is not Markov, it is *partially observable*. In partially observable systems predictions can depend arbitrarily on the entire history. We focus on the partially observable case.

## 2. Learning to Make Predictions

In this work we assume that, as in Three Card Monte, though the agent may live in a complex environment, it has only a small set of important predictions to make. These predictions could have been identified as important by a designer, or by some other learning process. We do not address the problem of identifying which predictions *should* be made, but rather focus on the problem of learning to make predictions, once they are identified. In general, we imagine that we are given some finite set $\mathcal{T}^I = \{t_1, t_2, \ldots, t_m\}$ of *tests of interest* for which we would like our model to make accurate predictions. Here the term "test" should be construed broadly, possibly including abstract tests in addition to raw sequences of actions and observations. The tests of interest are the future events the model should predict. For instance, in the Three Card Monte problem, in order to perform well the agent must predict whether it will see the ace when it flips over each card. So it will have three one-step tests of interest: *flip1 ace*, *flip2 ace*, and *flip3 ace* (representing the future events where the agent flips over card 1, 2, and 3, respectively, and sees the ace). If the agent can learn to maintain the probability of these events over time, it can win the game.

As such, the general problem is to learn a function $\phi : \mathcal{H} \to [0,1]^m$ where

$$\phi(h) \overset{\text{def}}{=} \langle p(t_1 \mid h), p(t_2 \mid h), \ldots, p(t_m \mid h) \rangle, \tag{2}$$

that is, a function from histories to the predictions for the test of interest (which we will refer to as the *predictions of interest*) at that history. Note that the output of $\phi$ is not necessarily a probability distribution. The tests of interest may be selected arbitrarily and therefore need not represent mutually exclusive or exhaustive events. We will call a particular vector of predictions for the tests of interest a *prediction profile*.

**Definition 6.** We call $\phi(h)$ the *prediction profile* at history $h$.

We now describe two existing general approaches to learning $\phi$: learning a direct function from history to predictions (most common in the Markov case), and learning a fully generative model that maintains a finite-dimensional summary of history (common in the partially observable case). Both have strengths and weaknesses as approaches to learning $\phi$. Section 2.3 will contrast these with our approach, which combines some of the strengths of both approaches.

### 2.1 Direct Function Approximation

When the system is Markov, learning $\phi$ is conceptually straightforward; essentially it is a problem of learning a function from observation ("state") to predictions. Rather than learning $\phi$ which takes histories as input, one can instead learn a function $\phi_{Markov} : \mathcal{O} \to [0,1]^m$, which maps an observation to the predictions for the tests of interest resulting from *all* histories that end in that observation. Note that, as an immediate consequence, in discrete Markov systems there is a finite number of *distinct* prediction profiles. In fact, there can be no more distinct prediction profiles than there are observations.

When the number of observations and the number of tests of interest are small enough, $\phi_{Markov}$ can be represented as a $|\mathcal{O}| \times |\mathcal{T}^I|$ look-up table, and the entries estimated using sample averages[1]:

$$\hat{p}(t_i \mid o) = \frac{\text{\# times } t \text{ succeeds from histories ending in } o}{\text{\# times } acts(t) \text{ taken from histories ending in } o}. \tag{3}$$

The main challenge of learning Markov models arises when the number of observations is very large. Then it becomes necessary to generalize across observations, using data gathered about one observation to learn about many others. Specifically, one may be able to exploit the fact that some observations will be associated with very similar (or identical) prediction profiles (that is, the same predictions for the tests of interest) and share data amongst them. Restricting a model's attention to only a few predictions can afford more generalization, which is why learning a partial model can be beneficial in the Markov setting.

Even when the system is partially observable, one can still attempt to learn $\phi$ directly, typically by performing some sort of regression over a set of features of *entire histories*. For instance, U-Tree (McCallum, 1995) takes a set of history features and learns a decision tree that attempts to distinguish histories that result in different expected asymptotic return under optimal behavior. Wolfe and Barto (2006) apply a U-Tree-like algorithm but rather than restricting the model to predicting future rewards, they learn to make predictions about some pre-selected set of features of the next observation (a special case of the more general concept of tests of interest). Dinculescu and Precup (2010) learn the expected value of a given feature of the future as a direct function of a given real-valued feature of history by clustering futures and histories that have similar associated values.

Because they directly approximate $\phi$ these types of models *only* make predictions for $\mathcal{T}^I$ and are therefore non-generative (and therefore able, for instance, to avoid falling into the trap of predicting the dealer's decisions in Three Card Monte). Though this approach has demonstrated promise, it also faces a clear pragmatic challenge, especially in the partially observable setting: feature selection. Because $\phi$ is a function of history, an ever-expanding sequence of actions and observations, finding a reasonable set of compactly represented features that collectively capture all of the history information needed to make the predictions of interest is a significant challenge. In a sense, even in the partially observable setting, this type of approach takes only a small step away from the Markov case. It still requires a good idea *a priori* of what information should be extracted from history (in the form of features) in order to make the predictions of interest.

---

1. Bowling, McCracken, James, Neufeld, and Wilkinson (2006) showed that this estimator is unbiased only in the case where the data is collected using a *blind* policy, in which action selection does not depend on the history of observations and provided an alternative estimator that is unbiased for all policies. For simplicity's sake, however, we will assume throughout that the data gathering policy is blind.

## 2.2 Generative Models

If one does not have a good idea *a priori* of what features should be extracted from history to make accurate predictions, one faces the additional challenge of *learning* to summarize the relevant information from history in a compact sufficient statistic.

There exist methods that learn from training data to maintain a finite-dimensional statistic of history from which *any* prediction can be computed. In analogy to the Markov case, this statistic is called the *state vector*. Clearly any model that can maintain state can be used to compute $\phi$ (since it can make *all* predictions). We briefly mention two examples of this approach that are particularly relevant to the development and analysis of our method.

**POMDPs** By far the most popular representation for models of partially observable systems is the partially observable Markov decision process (POMDP) (Monahan, 1982). A POMDP posits an underlying MDP (Puterman, 1994) with a set $\mathcal{S}$ of *hidden states* that the agent never observes. At any given time-step $i$, the system is in some particular hidden state $s_{i-1} \in \mathcal{S}$ (unknown to the agent). The agent takes some action $a_i \in \mathcal{A}$ and the system transitions to the next state $s_i$ according to the transition probability $\Pr(s_i \mid s_{i-1}, a_i)$. An observation $o_i \in \mathcal{O}$ is then emitted according to a probability distribution that in general may depend upon $s_{i-1}$, $a_i$, and $s_i$: $\Pr(o_i \mid s_{i-1}, a_i, s_i)$.

Because the agent does not observe the hidden states, it cannot know which hidden state the system is in at any given moment. The agent *can* however maintain a probability distribution that represents the agent's current *beliefs* about the hidden state. This probability distribution is called the *belief state*. If the belief state associated with history $h$ is known, then it is straightforward to compute the prediction of any test $t$:

$$p(t \mid h) = \sum_{s \in \mathcal{S}} \Pr(s \mid h) \Pr(t \mid s),$$

where $\Pr(t \mid s)$ can be computed using the transition and observation emission probabilities.

The belief state is a finite summary of history from which any prediction about the future can be computed. So, the belief state is the state vector for a POMDP. Given the transition probabilities and the observation emission probabilities, it is possible to maintain the belief state over time using Bayes' rule. If at the current history $h$ one knows $\Pr(s \mid h)$ for all hidden states $s$ and the agent takes action $a$ and observes observation $o$, then one can compute the probability of any hidden state $s$ at the new history:

$$\Pr(s \mid hao) = \frac{\sum_{s' \in \mathcal{S}} \Pr(s' \mid h) \Pr(s \mid s', a_i) \Pr(o_i \mid s', a_i, s)}{\sum_{s'' \in \mathcal{S}} \sum_{s' \in \mathcal{S}} \Pr(s' \mid h) \Pr(s'' \mid s', a_i) \Pr(o_i \mid s', a_i, s'')}. \tag{4}$$

The parameters of a POMDP that must be learned in order to be able to maintain state are the transition probabilities and the observation emission probabilities. Given these parameters, the belief state corresponding to any given history can be recursively computed and the model can thereby make any prediction at any history. POMDP parameters are typically learned using the Expectation Maximization (EM) algorithm (Baum et al., 1970). Given some training data and the number of actions, observations, and hidden states as input, EM essentially performs gradient ascent to find transition and emission distributions that (locally) maximize the likelihood of the provided data.

**PSRs** Another more recently introduced modeling representation is the predictive state representation (PSR) (Littman et al., 2002). Instead of hidden states, PSRs are defined more directly in terms of the system dynamics matrix (described in Section 1.2.2). Specifically, PSRs find a set of *core tests Q* whose corresponding columns in the system dynamics matrix form a basis. Recall that the system dynamics matrix often has finite rank (for instance, the matrix associated with any POMDP with finite hidden states has finite linear dimension) and thus $Q$ is finite for many systems of interest. Since the predictions of $Q$ are a basis, the prediction for *any* other test at some history can be computed as a linear combination of the predictions of $Q$ at that history.

The vector of predictions for $Q$ is called the *predictive state*. While the belief state was the state vector for POMDPs, the predictive state is the state vector for PSRs. It can also be maintained by application of Bayes' rule. Specifically, if at some history $h$, $p(q \mid h)$ is known for all core tests $q$ and the agent takes some action $a \in \mathcal{A}$ and observes some observation $o \in \mathcal{O}$, then one can compute the prediction of any core test $q$ at the new history:

$$p(q \mid hao) = \frac{p(aoq \mid h)}{p(ao \mid h)} = \frac{\sum_{q' \in Q} p(q' \mid h) m_{aoq}(q')}{\sum_{q' \in Q} p(q' \mid h) m_{ao}(q')}, \tag{5}$$

where $m_{aoq}(q')$ is the coefficient of $p(q' \mid h)$ in the linear combination that computes the prediction $p(aoq \mid h)$.

So, given a set of core tests, the parameters of a PSR that must be learned in order to maintain state are the coefficients $m_{ao}$ for every action $a$ and observation $o$ and the coefficients $m_{aoq}$ for every action $a$, observation $o$, and core tests $q$. Given these parameters the predictive state at any given history can be recursively computed and used to make any prediction about the future. PSRs are learned by directly estimating the system dynamics matrix (James & Singh, 2004; Wolfe, James, & Singh, 2005) or, more recently, some sub-matrix or derived matrix thereof (Boots, Siddiqi, & Gordon, 2010, 2011) using sample averages in the training data. The estimated matrix is used to find a set of core tests and the parameters are then estimated using linear regression.

Note that both of these types of models are *inherently* generative. They both rely upon the maintenance of the state vector in order to make predictions and, as can be seen in Equations 4 and 5, the state update equations of these models *rely upon access to one-step predictions* to perform the Bayesian update. As such, unlike the direct function approximation approach, one cannot simply choose a set of predictions for the model to make. These models by necessity make *all* predictions.

There are many reasons to desire a complete, generative model. Because it makes all possible predictions, such a model can be used to sample possible future trajectories which is a useful capability for planning. A generative model is also, by definition, very flexible about what predictions it can be used to make. On the other hand, in many cases a complete, generative model may be difficult to obtain. Both PSR and POMDP training methods scale very poorly with the linear dimension of the system being learned. The linear dimension lower-bounds the number of hidden states needed to represent a system as a POMDP and is precisely the number of core tests needed to represent it as a PSR. The learning methods for POMDPs and PSRs are rarely successfully applied to systems with a linear dimension of more than a few hundred (though the work of Boots et al. is pushing these limits further). Most systems of interest will have several orders of magnitude higher linear dimension.

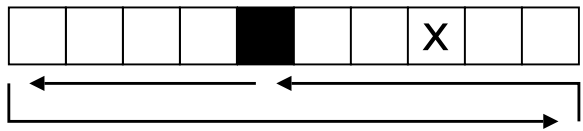

Figure 2: Size 10 1D Ball Bounce

Furthermore, a complete, generative model is overkill for the problem at hand. Recall that we do not seek to make *all* predictions; we are focused on making some particularly important predictions $\mathcal{T}^I$. Even in problems where learning to make all predictions might be intractable, it should still be possible to make *some* simple but important predictions.

### 2.2.1 Abstract Generative Models

As discussed earlier, when there is a restricted set of tests of interest, the learning problem can often be simplified by ignoring irrelevant details through abstraction. Of course, having an abstraction does not solve the problem of partial observability. What is typically done is to apply the abstraction to the training data, discarding the irrelevant details (as we did in the Three Card Monte example) and then to apply model learning methods like the ones described above to the *abstract* data set. Just as in the Markov setting, in some cases observation abstraction can greatly simplify the learning problem (certainly learning about only the cards in Three Card Monte is easier than learning about the cards *and* the crowd and weather and so on).

Ignoring details irrelevant to making the predictions of interest is intuitive and can significantly simplify the learning problem. On the other hand, because they are generative models, an abstract POMDP or PSR will still make *all abstract predictions*. This typically includes predictions other than those that are directly of interest. If these extra predictions require a complex model, even an abstract generative model can be intractible to learn. This is true of the Three Card Monte example (where a generative model ends up modeling the dealer as well as the cards). The following is another simple example of this phenomenon.

**Example.** Consider the uncontrolled system pictured in Figure 2, called the "1D Ball Bounce" system. The agent observes a strip of pixels that can be black or white. The black pixel represents the position of a ball that moves around on the strip. The ball has a current direction and every time-step it moves one pixel in that direction. Whenever it reaches an edge pixel, its current direction changes to move away from the edge. In Figure 3(a) a complete POMDP model of a 10 pixel version of this system is pictured. If there are $k$ pixels, the POMDP has $2k - 2$ hidden states (because the ball can have one of 2 possible directions in one of $k$ possible positions, except the two ends, where there is only one possible direction).

Now say the agent wishes only to predict whether the ball will be in the position marked with the $x$ in the next time step. Clearly this prediction can be made by only paying attention to the immediate neighborhood of the $x$. The details of what happens to the ball while it is far away do not matter for making these predictions. So, one could apply an abstraction that lumps together all observations in which the neighborhood about $x$ looks the same. The problem is that an abstract generative model of this system makes predictions not only about $x$, but also about the pixels surrounding $x$. Specifically, the model still makes predictions about whether the ball will enter the neighborhood in the near future. This of course depends on how long it has been since the ball left the neighborhood. So, the POMDP model of the abstract system (pictured in Figure 3(b)) has *exactly the same state diagram as the original system*, though its observations have changed to reflect the abstraction. The abstract system and the primitive system have the same linear dimension.

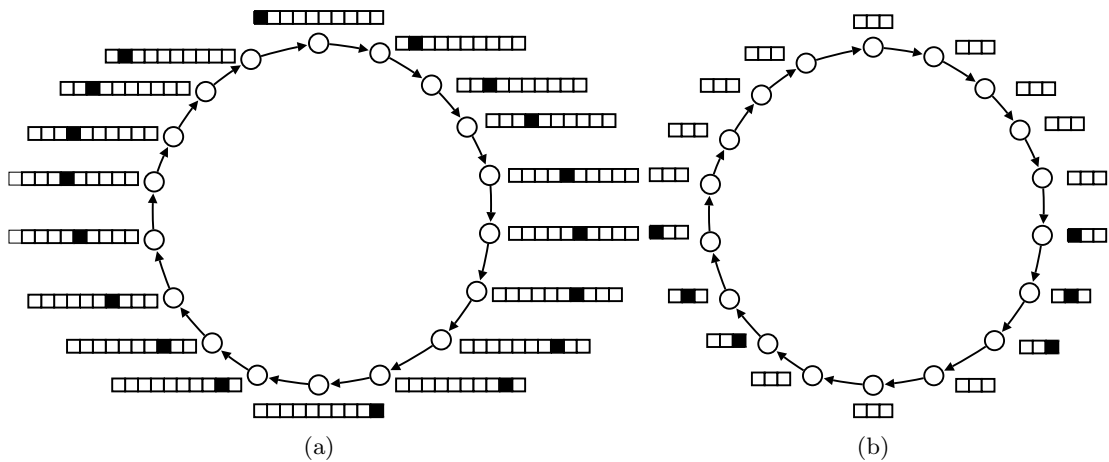

Figure 3: POMDP model of the size 10 1D Ball Bounce system (a) and of the abstracted 1D Ball Bounce system (b).

In order to make predictions about $x$, one must condition on information about the pixels surrounding $x$. Consequently, a generative model also makes predictions about these pixels. Counterintuitively, the abstract model's complexity is mainly devoted to making predictions other than the predictions of interest. In general, while learning an abstract model can drastically simplify the learning problem by ignoring irrelevant details, an abstract generative model still learns to make predictions about any details that *are* relevant, even if they are not directly of interest.

## 2.3 Prediction Profile Models

The contribution of this paper, *prediction profile models*, seek to combine the main strengths of the two model-learning approaches discussed above. As with a direct approximation of $\phi$, a prediction profile model will *only* make the predictions of interest, and no others. As such, it can be far simpler than a generative model, which will typically make many extraneous predictions. However, the learning method for prediction profile models will not require a set of history features to be given *a priori*. By leveraging existing generative model learning methods, prediction profile models learn to maintain the state information necessary for making the predictions of interest.

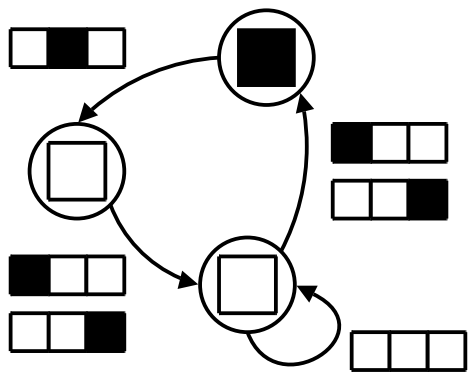

Figure 4: Prediction profile model for the 1D Ball Bounce system

A typical model learns to make predictions about future observations emitted by the system. The main idea behind prediction profile models is to instead model the *values of the predictions themselves* as they change over time, conditioned on both the actions chosen by the agent and the observations emitted by the system.

We have already seen an example of this in Three Card Monte. The prediction profile model (shown in Figure 1) takes observations of the dealer's behavior as *input* and outputs predictions for the tests of interest. It does not predict the dealer's behavior, but it takes it into account when updating the predictions of interest. Recall that, though the Three Card Monte system can be arbitrarily complicated (depending on the dealer), this prediction profile system has three states, regardless of the dealer's decision making process.

Another example is shown in Figure 4. This is the prediction profile system for the 1D Ball Bounce system (Figure 2), where the model must predict whether the ball will enter position $x$ in the next time-step. Each state of the prediction profile model is labeled with a prediction for pixel $x$ (white or black). The transitions are labeled with observations of the 3-pixel neighborhood centered on position $x$. In this case the transitions capture the ball entering the neighborhood, moving to position $x$, leaving the neighborhood, staying away for some undetermined amount of time, and returning again. Recall that a POMDP model of this system has $2k - 2$ hidden states, where $k$ is the number of pixels, even after ignoring all pixels irrelevant to making predictions about pixel $x$. By contrast, the prediction profile model always has three states, regardless of the number of pixels.

The next section will formally describe prediction profile models as models of a dynamical system that results from a transformation of the original system. Subsequent sections will discuss how to learn prediction profile models from data (by converting data from the original system into data from the transformed system and learning a model from the converted data set) and present results that help to characterize the conditions under which prediction profile models are best applied.

## 3. The Prediction Profile System

We now formally describe a theoretical dynamical system, defined in terms of *both* the dynamics of the original system and the given tests of interest. We call this constructed system the *prediction profile system*. A *prediction profile model*, which it is our goal to construct, is a model of the prediction profile system (that is, the system is an ideal, theoretical construct, a model may be imperfect, approximate, etc.). As such, our analysis of the problem of learning a prediction profile model will depend a great deal on understanding properties of the prediction profile system.

In this paper we make the restrictive assumption that, as in the Markov case, there is a finite number of distinct prediction profiles (that is, the predictions of interest take on only a finite number of distinct values). This is certainly not true of all partially observable systems and all sets of tests of interest, though it is true in many interesting examples. Formally, this assumption requires that $\phi$ map histories to a finite set of prediction profiles:

**Assumption 7.** *Assume there exists a finite set of prediction profiles $P = \{\rho_1, \rho_2, \ldots, \rho_k\} \subset [0,1]^m$ such that for every history $h$, $\phi(h) \in P$.*

This assumption allows the definition of the *prediction profile system* (or *PP* for short) as a discrete dynamical system that captures the sequence of prediction profiles over time, given an action observation sequence. The prediction profile system's actions, observations, and dynamics are defined in terms of quantities associated with the original system:

**Definition 8.** *The prediction profile system is defined by a set of observations, a set of actions, and a rule governing its dynamics.*

1. Observations: The set of prediction profile observations, $\mathcal{O}_{PP}$, is defined to be the set of distinct prediction profiles. That is, $\mathcal{O}_{PP} \overset{\text{def}}{=} P = \{\rho_1, \ldots, \rho_k\}$.

2. Actions: The set of prediction profile actions, $\mathcal{A}_{PP}$, is defined to be the set of action-observation pairs in the original system. That is, $\mathcal{A}_{PP} \overset{\text{def}}{=} \mathcal{A} \times \mathcal{O}$.

3. Dynamics: The dynamics of the prediction profile system are deterministically governed by $\phi$. At any prediction profile history, $\langle a^1, o^1\rangle\rho^1\langle a^2, o^2\rangle\rho^2\ldots\langle a^j, o^j\rangle\rho^j$, and for any next $PP$-action, $\langle a^{j+1}, o^{j+1}\rangle$, the prediction profile system deterministically emits the $PP$-observation $\phi(a^1 o^1 a^2 o^2 \ldots a^j o^j a^{j+1} o^{j+1})$.

We now present some key facts about the prediction profile system. Specifically, it will be noted that the prediction profile system is always deterministic. Also, though the prediction profile system may be Markov (as it is in the Three Card Monte example), in general it is partially observable.

**Proposition 9.** *Even if the original system is stochastic, the prediction profile system is always deterministic.*

*Proof.* This follows immediately from the definition: every history corresponds to exactly one prediction profile. So a $PP$-history (action-observation-profile sequence) and a $PP$-action (action-observation pair) fully determine the next $PP$-observation (prediction profile). The stochastic observations in the original system have been folded into the unmodeled *actions* of the prediction profile system. □

**Proposition 10.** *If the original system is Markov, the prediction profile system is Markov.*

*Proof.* By definition, if the original system is Markov the prediction profile at any time step depends only on the most recent observation. So, if at time step $t$, the current profile is $\rho_t$, the agent takes action $a_{t+1}$ and observes observation $o_{t+1}$, the next profile is simply $\rho_{t+1} = \phi_{Markov}(o_{t+1})$. So, in fact, when the original system is Markov, the prediction profile system satisfies an even stronger condition: the next $PP$-observation is *fully* determined by the $PP$-action and has no dependence on history whatsoever (including the most recent $PP$-observation). □

**Proposition 11.** *Even if the original system is partially observable, the prediction profile system may be Markov.*

*Proof.* Consider the Three Card Monte example. The original system is clearly non-Markov (the most recent observation, that is the dealer's most recent swap, tells one very little about the location of the ace). However, the prediction profile system for the tests of interest regarding the location of the special card (pictured in Figure 1) *is* Markov. The next profile is fully determined by the current profile and the $PP$-action. □

In general, however, the $PP$ system may be partially observable. Though in the Three Card Monte example the current prediction profile and the next action-observation pair together fully determine the next prediction profile, in general the next prediction profile is determined by the *history* of action-observation pairs (and prediction profiles).

**Proposition 12.** *The prediction profile system may be partially observable.*

*Proof.* Recall the 1D Ball Bounce example. The corresponding prediction profile system is shown in Figure 4. Note that two distinct states in the update graph are associated with the *same* prediction profile (pixel $x$ will be white). Given only the current prediction profile (pixel $x$ will be white) and the $PP$-action (observe the ball in a neighboring pixel on the left or right), one cannot determine whether the ball is entering or leaving the neighborhood, and thus cannot uniquely determine the next profile. This prediction profile system is partially observable. □

So, in general, the prediction profile system is a deterministic, partially-observable dynamical system. A model of the prediction profile system can *only* be used to make the predictions of interest. As such, if one wishes to use a prediction profile model as a *generative* model, one must select the tests of interest carefully. For instance:

**Proposition 13.** *If the tests of interest include the set of one-step primitive tests, that is if $\{ao \mid a \in \mathcal{A}, o \in \mathcal{O}\} \subseteq \mathcal{T}^I$, then a model of the prediction profile system can be used as a generative model of the original system.*

*Proof.* This follows immediately from the definition of generative model. □

While in this special case a prediction profile model can be a complete, generative model of the system, it will be shown in Section 5 that if one desires a generative model, it is essentially never preferable to learn a prediction profile model over a more traditional representation. A prediction profile model is *best* applied when it is relatively simple to make and maintain the predictions of interest in comparison to making *all* predictions. In general, a prediction profile model conditions on the observations, but it does not necessarily *predict* the next observation. As such, a model of the prediction profile system cannot typically be used for the purposes of model-based planning/control like a generative model could. The experiments in Section 6 will demonstrate that the output of prediction profile models *can*, however, be useful for model-free control methods.

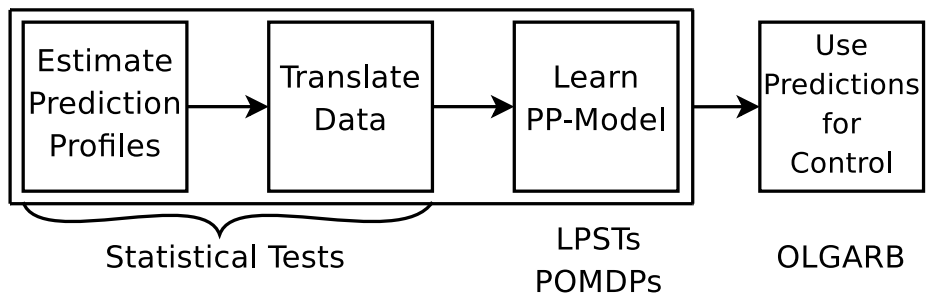

Figure 5: Flow of the algorithm.

## 4. Learning a Prediction Profile Model

The definition of the prediction profile system straightforwardly suggests a method for learning prediction profile models (estimate the prediction profiles, and learn a model of their dynamics using a standard model-learning technique). This section will present such a learning algorithm, discussing some of the main practical challenges that arise.

Let $S$ be a training data set of trajectories of experience with the original system (action-observation sequences) and let $\mathcal{T}^I = \{t_1, t_2, \ldots, t_k\}$ be the set of tests of interest. The algorithm presented in this section will learn a model of the prediction profile system from the data $S$. The algorithm has three main steps (pictured in Figure 5). First the training data is used to estimate the prediction profiles (both the number of unique profiles and their values). Next, the learned set of prediction profiles is used to translate the training data into trajectories of experience with the prediction profile system. Finally, any applicable model learning method can be trained on the transformed data to learn a model of the prediction profile system. Ultimately, in our experiments, the learned prediction profile models will be evaluated by how useful their predictions are as features for control.

### 4.1 Estimating the Prediction Profiles

Given $S$ and $\mathcal{T}^I$, the first step of learning a prediction profile model is to determine how many distinct prediction profiles there are, as well as their values. The estimated prediction for a test of interest $t$ at a history $h$ is:

$$\hat{p}(t \mid h) = \frac{\#\text{ times } t \text{ succeeds from } h}{\#\text{ times } acts(t) \text{ taken from } h}. \tag{6}$$

One could, at this point, directly estimate $\phi$ by letting $\hat{\phi}(h) \stackrel{\text{def}}{=} \langle \hat{p}(t_1 \mid h), \hat{p}(t_2 \mid h), \ldots, \hat{p}(t_k \mid h)\rangle$. Of course, due to sampling error, it is unlikely that *any* of these estimated profiles will be exactly the same, even if the true underlying prediction profiles are identical. So, to estimate the number of distinct underlying profiles, statistical tests will be used to find histories that have *significantly* different prediction profiles.

To compare the profiles of two histories, a likelihood-ratio test of homogeneity is performed on the counts for each test of interest in the two histories. If the statistical test associated with any test of interest rejects the null hypothesis that the prediction is the same in both histories, then the two histories have different prediction profiles.

In order to find the set of distinct prediction profiles, we greedily cluster the estimated prediction profiles. Specifically, an initially empty set of exemplar histories is maintained. The algorithm searches over all histories in the agent's experience, comparing each history's estimated profile to the exemplar histories' estimated profiles. If the candidate history's profile is significantly different from the profiles of *all* exemplar histories, the candidate is added as a new exemplar. In the end, the estimated profiles corresponding to the exemplar histories are used as the set of prediction profiles. In order to obtain the best estimates possible, the search is ordered so as to prioritize histories with lots of associated data.

The prediction profile estimation procedure has two main sources of complexity. The first is the sample complexity of estimating the prediction profiles. It can take a great deal of exploration to see each history enough times to obtain good statistics, especially if the number of actions and observations is large. This issue could be addressed by adding generalization to the estimation procedure, so that data from one sample trajectory could improve the estimates of many similar histories. In one of the experiments in Section 6, observation abstraction will be employed as a simple form of generalization. The second bottleneck is the computational complexity of searching for prediction profiles, as this involves exhaustively enumerating all histories in the agent's experience. It would be valuable to develop heuristics to identify the histories most likely to provide new profiles, in order to avoid searching over all histories. In the experiments in Section 6, a simple heuristic of limiting the search to short histories is employed. Long histories will tend to have less associated data, and will therefore be less likely to provide distinguishably new profiles.

## 4.2 Generating Prediction Profile Trajectories

Having generated a finite set of distinct prediction profiles, the next step is to translate the agent's experience into sequences of action-observation pairs and prediction profiles. These trajectories will be used to train a model of the prediction profile system.

The process of translating an action-observation sequence $s$ into a prediction profile trajectory $s'$ is straightforward and, apart from a few practical concerns, follows directly from Definition 8. Recall that, for an action-observation sequence $s = a_1 o_1 a_2 o_2 \ldots a_k o_k$, the corresponding $PP$-action sequence is $\langle a_1, o_1 \rangle \langle a_2, o_2 \rangle \ldots \langle a_k, o_k \rangle$. The corresponding sequence of profiles is $\phi(a_1 o_1) \phi(a_1 o_1 a_2 o_2) \ldots \phi(a_1 o_1 \ldots a_k o_k)$. Thus, in principle, every primitive action-observation sequence can be translated into an action-observation-profile sequence.

Of course $\phi$ is not available to generate the sequence of prediction profiles. So, it is necessary to use an approximation $\hat{\phi}$, generated from the training data. Specifically, the estimated predictions for the tests of interest at each history $h$ (computed using Equation 6) are compared, using statistical tests, to the set of distinct estimated prediction profiles from Section 4.1. If there is only one estimated profile $\hat{\rho}$ that is not statistically significantly different from the estimated predictions at $h$, then let $\hat{\phi}(h) = \hat{\rho}$.

Given sufficient data, the statistical tests will uniquely identify the correct match with high probability. In practice, however, some histories will not have very much associated data. It is possible in such a case for the test of homogeneity to fail to reject the null hypothesis for two or more profiles. This indicates that there is not enough data to distinguish between multiple possible matches. In the experiments in Section 6, two different heuristic strategies for handling this situation are employed. The first strategy lets $\hat{\phi}(h)$ be the matching profile that has the smallest empirical KL-Divergence from the estimated predictions (summed over all tests of interest). This is a heuristic choice that may lead to noise in the prediction profile labeling, which could in turn affect the accuracy of the learned model. The second strategy is to simply cut off any trajectory at the point where multiple matches occur, rather than risk assigning an incorrect labeling. This ensures that labels only appear in the prediction profile trajectories if there is a reasonable level of confidence in their correctness. However, it is wasteful to throw out training data in this way.

## 4.3 Learning a Prediction Profile Model

The translation step produces a set $S'$ of trajectories of interaction with the prediction profile system. Recall that the prediction profile system is a deterministic, partially observable, discrete dynamical system and these trajectories can be used to train a model of the prediction profile system using, in principle, any applicable model-learning method.

There is an issue faced by models of the prediction profile system that is not present in the usual discrete dynamical systems modeling setting. While the prediction profile labels are present in the training data, when actually *using* the model they are not available. Say the current history is $h$, and an action $a_1$ is taken and an observation $o_1$ is emitted. Together, this action-observation pair constitutes a $PP$-action. Being a model of the prediction profile system, a prediction profile model can identify the next profile, $\rho$. This profile can be used to compute predictions $p(t \mid ha_1o_1)$ for the tests of interest at the history $ha_1o_1$. Now another action $a_2$ and observation $o_2$ occur. It is now necessary to update the PP-model's state in order to obtain the next prediction profile. A typical dynamical systems model makes predictions about the next observation, but is then able to update its state with the *actual* observation that occurred. A prediction profile model's observations are prediction profiles themselves, which are not observable when interacting with the world. As such, the prediction profile model will update its state with prediction profile *it itself predicted* ($\rho$). Once updated, the prediction profile model can obtain the profile that follows $\langle a_2, o_2 \rangle$ which gives the predictions for the tests of interest at the new history $ha_1o_1a_2o_2$.

If the prediction profile model is a *perfect* model of the prediction profile system, this poses no problems. Because the prediction profile system is deterministic, there is no need to observe the true prediction profile label; it is fully determined by the history. In practice, of course, the model will be imperfect and different modeling representations will require different considerations when performing the two functions of providing predictions for the tests of interest, and providing a profile for the sake of updating the model.

### 4.3.1 PP-POMDPs

Since the prediction profile system is partially observable it is natural to model it using a POMDP. Unfortunately, even when the training data is from a deterministic sys-tem, POMDP training using the EM algorithm will generally not provide a deterministic POMDP. Thus, at any given history, a learned POMDP model of the prediction profile system (PP-POMDP) will provide a *distribution* over prediction profiles instead of deterministically providing the one profile associated with that history. The implementation used in Section 6 simply takes the *most likely* profile from the distribution to be the profile associated with the history and uses it to make predictions for the tests of interest, as well as to update the POMDP model.

### 4.3.2 PP-LPSTs

Another natural choice of representation for a prediction profile model is a looping predictive suffix tree (LPST) (Holmes & Isbell, 2006). LPSTs are specialized to deterministic, partially observable systems. As such, they could not be used to model the original system (which is assumed to be stochastic in general), but they do apply to the prediction profile system (and they do not have to be determinized like a POMDP).

Briefly, an LPST captures the parts of recent history relevant to predicting the next observation. Every node in the tree corresponds to an action-observation pair. A node may be a leaf, may have children, or it may loop to one of its ancestors. Every leaf of the tree corresponds to a history suffix that has a deterministic prediction of an observation for every action. In order to predict the next observation from a particular history, one reads the history in reverse order, following the corresponding links on the tree until a leaf is reached, which gives the prediction. Holmes and Isbell provide a learning algorithm that, under certain conditions on the training data, is guaranteed to produce an optimal tree. The reader is referred to the work of Holmes and Isbell (2006) for details.

One weakness of LPSTs, however, is that they fail to make a prediction for the next observation if the current history does not lead to a leaf node in the tree (or if the leaf node reached does not have a prediction for the action being queried). This typically occurs when some history suffixes do not occur in the training data but do occur while using the model. For a PP-LPST, this can mean that in some histories the model cannot uniquely determine the corresponding prediction profile. When this happens the implementation used in Section 6 simply finds the longest suffix of the current history that *does* occur in the data. This suffix will be associated with multiple prediction profiles (otherwise the LPST would have provided a prediction). To make predictions for the tests of interest, the model provides the average prediction over this set of profiles. The profile used to update the model is picked out of the set uniformly randomly.

### 4.3.3 PP-PSRs

Applying PSR learning algorithms to prediction profile data poses a practical concern. Specifically, methods that attempt to estimate the system dynamics matrix (James & Singh, 2004; Wolfe et al., 2005) implicitly presume that every action sequence could in principle be taken from every history. If some action sequences can be taken from some histories but not from others, then the matrix will have undefined entries. This poses challenges to rank estimation (and, indeed, the very definition of the model representation). Unfortunately, this can be the case for the prediction profile system since $PP$-actions (action-observation pairs) are not completely under the agent's control; they are partly selected by the environ-ment itself. The recent spectral learning algorithms presented by Boots et al. (2010) may be able to side-step this issue, as they have more flexibility in selecting which predictions are estimated for use in the model-learning process, though we have not investigated this possibility in this work.

Note that, though our method for learning a prediction profile model involves standard model-learning methods for partially observable environments, the result is *not* a generative model of the original system. A prediction profile model is a generative model of the *prediction profile system* and, as such, cannot be used to make any predictions about the original system, other than the predictions of interest.

## 5. Complexity of the Prediction Profile System

The learning algorithm we have presented will be evaluated empirically in Section 6. First, however, we analyze the complexity of the prediction profile system in relation to the complexity of the original system. This will give some indication of how difficult it is to learn a prediction profile model and provide insight into when it is appropriate to learn a prediction profile model over a more typical generative model approach.

There are many factors that affect the complexity of learning a model. This section will largely focus on linear dimension as the measure of complexity, taking the view that, generally speaking, systems with lower linear dimension are easier to learn than systems with larger linear dimension. As discussed in Section 1.2.2, this is generally true for POMDPs, where the linear dimension lower-bounds the number of hidden states. So comparing the linear dimension of the prediction profile system to that of the original system can give some idea of whether it would be easier to learn a PP-POMDP or just to learn a standard POMDP of the original system. Of course, there are other model-learning methods for which other complexity measures would be more appropriate (for instance it is not known precisely how LPSTs interact with linear dimension). Extending some of these results to other measures of complexity may be an interesting topic of future investigation.

### 5.1 Linear Dimension Comparison

This section will discuss how the linear dimension of the prediction profile system relates to that of the original system. The first result is a "proof of concept" that simply states that there exist problems in which the prediction profile system is vastly more simple than the original system. In fact, such a problem has already been presented.

**Proposition 14.** *The prediction profile system can have linear dimension that is arbitrarily lower than that of the original system.*

*Proof.* Recall the Three Card Monte example. Thus far the domain has been described without describing the dealer's behavior. However, note that the prediction profile system for the tests of interest relating to the location of the special card (pictured in Figure 1) has a linear dimension of 3, *regardless of how the dealer's swaps are chosen.* If a very complex dealer is chosen, the original system will have high linear dimension, but the prediction profile system's linear dimension will remain constant. For instance, in the experiments in Section 6, the dealer chooses which cards to swap stochastically, but is more likely to choose the swap that has been selected the least often so far. Thus, in order to predict the dealer's next decision, one must count how many times each swap has been chosen in history and as a result the system effectively has infinite linear dimension. □

On the other hand, prediction profile models are not a panacea. The following results indicate that there are problems for which learning a prediction profile model would not be advisable over learning a standard generative model, in that the linear dimension of the prediction profile system can be far greater than that of the original system. Later in the section some special cases will be characterized where prediction profile models are likely to be useful. The next result shows that the linear dimension of the prediction profile model can be infinite when the original system has finite linear dimension, via a lower bound on linear dimension that is true of all deterministic dynamical systems.

**Proposition 15.** *For any deterministic dynamical system with actions $\mathcal{A}$, and observations $\mathcal{O}$, the linear dimension, $n \geq \frac{\log(|\mathcal{A}|-1)+\log(|\mathcal{O}|+1)}{\log|\mathcal{A}|}$.*

*Proof.* See Appendix A.1. □

Because Proposition 15 applies to all deterministic dynamical systems, it certainly applies to the prediction profile system. Though it is a very loose bound, the basic implication is that as the number of prediction profiles (the observations of $PP$) increases in comparison to the number of action-observation pairs (the actions of $PP$), the linear dimension of the prediction profile system necessarily increases. This bound also clearly illustrates the importance of the assumption that there is a finite number of distinct prediction profiles.

**Corollary 16.** *If there are infinitely many distinct prediction profiles, the prediction profile system has infinite linear dimension.*

*Proof.* Clearly $|\mathcal{A}_{PP}| = |\mathcal{A} \times \mathcal{O}|$ is finite so long as there are finitely many actions and observations. So, from the last result it follows immediately that as the number of distinct prediction profiles $|\mathcal{O}_{PP}|$ approaches infinity, then so must the linear dimension of the prediction profile system. □

Hence, so long as prediction profile models are represented using methods that rely on a finite linear dimension, it is critical that there be finitely many prediction profiles. Note that this is not a fundamental barrier, but a side effect of the representational choice. Model learning methods that are not as sensitive to linear dimension (such as those designed to model continuous dynamical systems) may be able to effectively capture systems with infinitely many prediction profiles.

One conclusion to be drawn from the last few results is that knowing the linear dimension of the original system does not, in itself, necessarily say much about the complexity of the prediction profile system. The prediction profile system may be far simpler or far more complex than the original system. Thus it may be more informative to turn to other factors when trying to characterize the complexity of the prediction profile system.

## 5.2 Bounding the Complexity of The Prediction Profile System

The results in the previous section do not take into account an obviously important aspect of the prediction profile system: the predictions it is asked to make. Some predictions of interest can be made very simply by keeping track of very little information. Other predictions will rely on a great deal of history information and will therefore require a more complex model. The next result identifies the "worst case" set of tests of interest for any system: the tests of interest whose corresponding prediction profile model has the highest linear dimension. Ultimately this section will present some (non-exhaustive) conditions under which the prediction profile system is likely to be simpler than the original system.

**Proposition 17.** *For a given system and set of tests of interest, the linear dimension of the corresponding prediction profile system is no greater than that of the prediction profile system associated with any set of core tests for the system (as described in Section 2.2).*

*Proof.* See Appendix A.2. □

With this worst case identified, one can immediately obtain bounds on how complex any prediction profile system can possibly be.

**Corollary 18.** *For any system and any set of tests of interest, the corresponding prediction profile system has linear dimension no greater than the number of distinct predictive states for the original system.*

*Proof.* The prediction profile system for a set of core tests $Q$ is a deterministic MDP where the observations are prediction profiles for $Q$ (that is, predictive states). That is, each state is associated with a unique prediction profile. The linear dimension of an MDP is never greater than the number of observations (Singh et al., 2004). Therefore, by the previous result the prediction profile system for any set of tests of interest can have linear dimension no greater than the number of predictive states. □

**Corollary 19.** *If the original system is a POMDP, the prediction profile system for any set of tests of interest has linear dimension no greater than the number of distinct belief states.*

*Proof.* This follows immediately from the previous result and the fact that the number of distinct predictive states is no greater than the number of distinct belief states (Littman et al., 2002). □

The bounds presented so far help explain why the prediction profile system can be more complex than the original system. However, because they are focused on the worst possible choice of tests of interest, they do little to illuminate when the opposite is true. A prediction profile model is at its most complex when it is asked to perform the same task as a generative model: keep track of as much information from history as is necessary to make *all possible predictions* (or equivalently, the predictive state or the belief state). These results indicate that, generally speaking, if one desires a generative model, standard approaches would be preferable to learning a prediction profile model.

On the other hand, our stated goal is *not* to learn a generative model, but instead to focus on some particular predictions that will hopefully be far simpler to make than *all* predictions. The examples we have seen make it clear that in some cases, some predictions can be made by a prediction profile model far simpler than a generative model of the original system. In general one might expect the prediction profile model to be simple when the predictions of interest rely on only a small amount of the state information required to maintain a generative model. The next bound aligns with this intuitive reasoning.

Essentially what this result points out is that often much of the hidden state information in a POMDP will be irrelevant to the predictions of interest. The linear dimension of the prediction profile system is bounded only by the number of distinct beliefs over the *relevant* parts of the hidden state, rather than the number of distinct beliefs states overall. The idea of the result is that if one can impose an abstraction over the *hidden states* of a POMDP (not the observations) that still allows the predictions of interest to be made accurately and that allows abstract belief states to be computed accurately, then the prediction profile system's linear dimension is bounded by the number of *abstract* belief states.

**Proposition 20.** *Consider a POMDP with hidden states $\mathcal{S}$, actions $\mathcal{A}$, and observations $\mathcal{O}$. Let $\mathcal{T}^I$ be the set of tests of interest. Let $a^i$ be the action taken at time-step $i$, $s^i$ be the hidden state reached after taking action $a^i$, and $o^i$ be the observation emitted by $s^i$. Now, consider* any *surjection $\sigma : \mathcal{S} \to \mathcal{S}^\sigma$ mapping hidden states to a set of abstract states with the following properties:*

1. *For any pair of primitive states $s_1, s_2 \in \mathcal{S}$, if $\sigma(s_1) = \sigma(s_2)$, then for any time-step $i$ and any test of interest $t \in \mathcal{T}^I$, $p(t \mid s^i = s_1) = p(t \mid s^i = s_2)$.*

2. *For any pair of primitive states $s_1, s_2 \in \mathcal{S}$, if $\sigma(s_1) = \sigma(s_2)$, then for any time-step $i$, abstract state $S \in \mathcal{S}^\sigma$, observation $o \in \mathcal{O}$, and action $a \in \mathcal{A}$,*

$$Pr(\sigma(s^{i+1}) = S \mid s^i = s_1, a^{i+1} = a, o^{i+1} = o) =$$
$$Pr(\sigma(s^{i+1}) = S \mid s^i = s_2, a^{i+1} = a, o^{i+1} = o).$$

*For any such $\sigma$, the prediction profile system for $\mathcal{T}^I$ has linear dimension no greater than the number of distinct beliefs over abstract states, $\mathcal{S}^\sigma$.*

*Proof.* See Appendix A.3 □

There are a few things to note about this result. First, a surjection $\sigma$ *always* exists that has properties 1 and 2. One can always define $\sigma : \mathcal{S} \to \mathcal{S}$ with $\sigma(s) \stackrel{\text{def}}{=} s$. This degenerate case trivially satisfies the requirements of Proposition 20 and recovers the bound given in Corollary 19. However, Proposition 20 applies to *all* surjections that satisfy the conditions. There must be a surjection that satisfies the conditions *and* results in the smallest number of beliefs over abstract states. Essentially, this is the one that ignores as much state information as possible while still allowing the predictions of interest to be made accurately and it is *this* surjection that most tightly bounds the complexity of the prediction profile system (even if $\sigma$ is not known).

Of course, there may still be a large or even infinite number of distinct beliefs, even over abstract states, so other factors must come into play to ensure a simple prediction profile system. Furthermore, this result does not characterize all settings in which the prediction profile system will be simple. That said, this result does support the intuition that the prediction profile system will tend to be simple when the predictions it is asked to make depend on small amounts of state information.

In order to build intuition about how this result relates to earlier examples, recall the Three Card Monte problem. In Three Card Monte there are two sources of hidden state: the ace's unobserved position and whatever hidden mechanism the dealer uses to make its decisions. Clearly the agent's predictions of interest depend only on the first part of the hidden state. So, in this case one can satisfy Property 1 with a surjection $\sigma$ that maps two hidden states to the same abstract state if the ace is in the same position, regardless of the dealer's state. Under this $\sigma$ there are only 3 abstract states (one for each possible position), even though there might be infinitely many true hidden states. Now, different states corresponding to the same ace position will have different distributions over the ace's next position; this distribution does, after all, depend upon the dealer's state. However, Property 2 is a statement about the distribution over the next abstract state *given* the observation that is emitted after entering the abstract state. If one knows the current abstract state and *observes* what the dealer does, the next abstract state is fully determined. So Property 2 holds as well. In fact, since the ace's position is known at the beginning of the game, this means the current abstract state is always known with absolute certainty, even though beliefs about the dealer's state will in general be uncertain. Hence, there are only 3 distinct beliefs about the abstract states (one for each state). As such, the prediction profile model's linear dimension is upper-bounded by 3, regardless of the dealer's complexity (and in this case the bound is met).

### 5.3 Bounding the Number of Prediction Profiles

The previous section describes some conditions under which the prediction profile system may have a lower linear dimension than the original system. Also of concern is the number of prediction profiles, and whether that number is finite. This section will briefly discuss some (non-exhaustive) cases in which the number of prediction profiles is bounded.

One case that has already been discussed is when the original system is Markov. In that case the number of prediction profiles is bounded by the number of observations (states). Of course, when the original system is Markov, there is little need to use prediction profile models. Another, similar case is when the system is partially observable, but completely deterministic (that is, the next observation is completely determined by history and the selected action). If the system is a deterministic POMDP then at any given history the current hidden state is known. As such, the number of belief states is bounded by the number of hidden states. Since there cannot be more prediction profiles than belief states, the number of prediction profiles are bounded as well.

One can move away from determinism in a few different ways. First, note that the key property of a deterministic POMDP is that the hidden state is fully determined by history. It is possible to satisfy this property even in stochastic systems, as long as one can uniquely determine the hidden state, *given* the observation that was emitted when arriving there. In that case, observations can be emitted stochastically, but the number of belief states (and the number of prediction profiles) is still bounded by the number of hidden states.

Another step away from determinism is a class of systems, introduced by Littman (1996), called Det-POMDPs. A Det-POMDP is a POMDP where the transition function and the observation function are both deterministic, but the initial state distribution may be stochastic. A Det-POMDP is *not* a deterministic dynamical system, as there is uncertainty about the hidden state. Because of this uncertainty, the system appears to emit observations stochastically. It is only the *underlying dynamics* that are deterministic. Littman showed that a Det-POMDP with $n$ hidden states and an initial state distribution with $m$ states in its support has at most $(n+1)^m - 1$ distinct belief states. So, this bounds the number of prediction profiles as well.

Finally, and most importantly, if the hidden state can be abstracted as in Proposition 20, then these properties only really need to hold for *abstract beliefs*. That is, the environment itself may be complex and stochastic in arbitrary ways, but if the *abstract* hidden state described in Proposition 20 is fully determined by history, then the number of prediction profiles is bounded by the number of abstract states (as was the case in Three Card Monte). Similarly, Det-POMDP-like properties can be imagined for abstract hidden states as well.

These cases by no means cover all situations where the number of prediction profiles can be bounded, but they do seem to indicate that the class of problems where the number of prediction profiles is finite is quite broad, and may contain many interesting examples.

## 6. Experiments

This section will empirically evaluate the prediction profile model learning procedure developed in Section 4. In each experiment an agent faces an environment for which a generative model would be a challenge to learn due to its high linear dimension. However, in each problem the agent could make good decisions if it could only have the predictions to a small number of important tests. A prediction profile model is learned for these important tests and the accuracy of the learned predictions is evaluated.

These experiments also demonstrate one possible use of prediction profile models (and partial models in general) for control. Because they are not generative, prediction profile models cannot typically be used directly by offline, model-based planning methods. However, their output may be useful for model-free methods of control. Specifically, in these experiments, the predictions made by the learned prediction profile models are provided as features to a *policy gradient* algorithm.

### 6.1 Predictive Features for Policy Gradient

Policy gradient methods (e.g., Williams, 1992; Baxter & Bartlett, 2000; Peters & Schaal, 2008) have been very successful as viable options for model-free control in partially observable domains. Though there are differences between various algorithms, the common thread is that they assume a parametric form for the agent's policy and then attempt to alter those parameters in the direction of the gradient with respect to expected average reward. These experiments will make use of Online GPOMDP with Average Reward Baseline (Weaver & Tao, 2001), or OLGARB (readers are referred to the original paper for details). OLGARB assumes there is some set of features of history, and that the agent's policy takes the parametric form:

$$\Pr(a \mid h; \vec{w}) = \frac{e^{\sum_i w_{i,a} f_i(h)}}{\sum_{a'} e^{\sum_i w_{i,a'} f_i(h)}}$$

where $f_i(h)$ is the $i$th feature and each parameter $w_{i,a}$ is a weight specific to the feature *and* the action being considered.

Typically the features used in policy gradient are features that can be directly read from history (e.g., features of the most recent few observations or the presence/absence of some event in history). It can be difficult to know *a priori* which historical features will be important for making good control decisions. In contrast, the idea in these experiments is to provide the values of some *predictions* as features. These *predictive features* have direct consequences for control, as they provide information about the effects of possible behaviors the agent might engage in. As such, it may be easier to select a set of predictive features that are likely to be informative about the optimal action to take (e.g., "Will the agent reach the goal state when it takes this action?" or "Will taking this action damage the agent?"). Furthermore, information may be expressed compactly in terms of a prediction that would be complex to specify purely in terms of past observations. As seen in the discussion of PSRs in Section 2.2, an arbitrary-length history can be fully captured by a finite set of short-term predictions. For these reasons it seems reasonable to speculate that predictive features, as maintained by a prediction profile model, may be particularly valuable to model-free control methods like policy gradient.

## 6.2 Experimental Setup

The learning algorithm will be applied to two example problems. In each problem prediction profile models are learned with various amounts of training data (using both LPSTs and POMDPs as the representation and using both strategies for dealing with multiple matches, as described in Section 4.3). The prediction accuracy of the models is evaluated, as well as how useful their predictions are as features for control. The training data is generated by executing a uniform random policy in the environment.

The free parameter of the learning algorithm is the significance value of the statistical tests, $\alpha$. Given the large number of contingency tests that will be performed on the same data set, which can compound the probability of a false negative, $\alpha$ should be set fairly low. In these experiments we use $\alpha = 0.00001$, though several reasonable values were tried with similar results. As discussed in Section 4, there will also be a maximum length of histories to consider during the search for prediction profiles. This cutoff allows the search to avoid considering long histories, as there are many long histories to search over and they are unlikely to provide new prediction profiles.

After a prediction profile model is learned, its predictions are evaluated as features for the policy gradient algorithm OLGARB. Specifically, for each test of interest $t$ the unit interval is split up into 10 equally-sized bins $b$ and a binary feature $f_{t,b}$ is provided that is 1 if the prediction of $t$ lies in bin $b$, and 0 otherwise. Also provided are binary features $f_o$, for each possible observation $o$. The feature $f_o = 1$ if $o$ is the most recent observation and 0, otherwise. The parameters of OLGARB, the learning rate and discount factor, are set to 0.01 and 0.95, respectively in all experiments.

To evaluate a prediction profile model OLGARB is run for 1,000,000 steps. The average reward obtained and the root mean squared error (RMSE) of the predictions for the tests of interest accrued by the model along the way are reported. Prediction performance is compared to that obtained by learning a POMDP on the training data and using it to make the predictions of interest. Because these problems are too complex to feasibly train a POMDP with the correct number of underlying states, 30-state POMDPs were used (stopping EM after a maximum of 50 iterations)[2]. Control performance is compared to that obtained by OLGARB using the predictions provided by a learned POMDP model as features, as well as OLGARB using the true predictions as features (the best the prediction profile model could hope to do), OLGARB using second-order Markov features (the two most recent observations, as well as the action between them) but no predictive features at all, and a hand-coded expert policy.

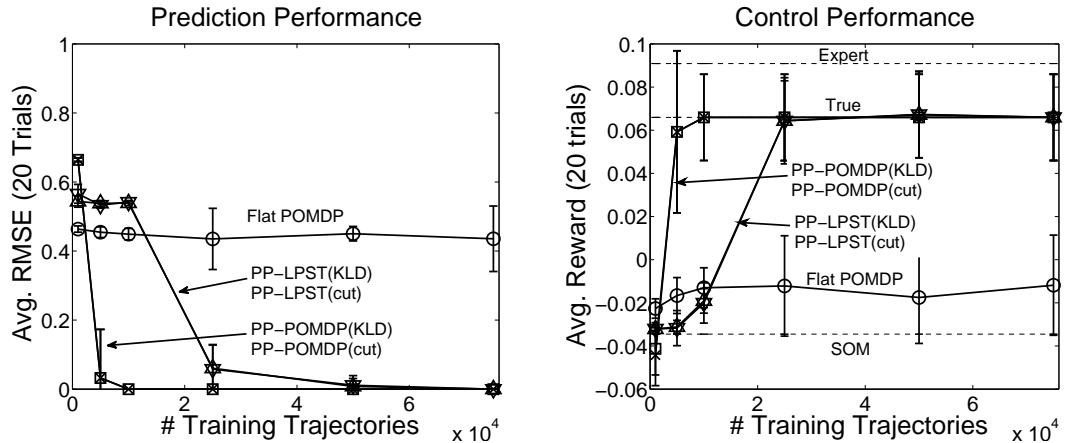

Figure 6: Results in the Three Card Monte domain.

### 6.3 Three Card Monte

The first domain is the Three Card Monte example. The agent is presented with three cards. Initially, the card in the middle (card 2) is the ace. The agent has four actions available to it: *watch*, *flip1*, *flip2*, and *flip3*. If the agent chooses a flip action, it observes whether the card it flipped over is the special card. If the agent chooses the *watch* action, the dealer can swap the positions of two cards, in which case the agent observes which two cards were swapped, or the dealer can ask for a guess. If the dealer has not asked for a guess, then *watch* results in 0 reward and any flip action results in -1 reward. If the dealer asks for a guess and the agent flips over the special card, the agent gets reward of 1. If the agent flips over one of the other two cards, or doesn't flip a card (by selecting *watch*), it gets reward of -1. The agent has three tests of interest, and they take the form *flipX ace*, for each card $X$ (that is, "If I flip card $X$, will I see the ace?").

As discussed previously, the complexity of this system is directly related to the complexity of the dealer's decision-making process. In this experiment, when the agent chooses "watch" the dealer swaps the pair of cards it has swapped the least so far with probability 0.5; with probability 0.4 it chooses uniformly amongst the other pairs of cards; otherwise it asks for a guess. Since the dealer is keeping a count of how many times each swap was made, the process governing its dynamics effectively has an infinite linear dimension. The prediction profile system, on the other hand, has only 3 states, regardless of the dealer's complexity (see Figure 1).

---

2. Similar results were obtained with 5, 10, 15, 20, and 25 states.

Training trajectories were of length 10. Figure 6 shows the results for various amounts of training data, averaged over 20 trials. Both PP-POMDPs and PP-LPSTs learned to make accurate predictions for the tests of interest, eventually achieving zero prediction error. In this case, PP-POMDPs did so using less data. This is likely because a POMDP model is more readily able to take advantage of the fact that the prediction profile system for Three Card Monte is Markov. As expected, the standard POMDP model was unable to accurately predict the tests of interest.

Also compared are the two different strategies for dealing with multiple matches discussed in Section 4.3. Recall that the first one (marked "KLD" in the graph) picks the matching profile with the smallest empirical KL-Divergence from the estimated predictions. The second (marked "cut" in the graph) simply cuts off the trajectory at the point of a multiple match to avoid any incorrect labels. In this problem these two strategies result in almost exactly the same performance. This is likely because the profiles in Three Card Monte are deterministic, and are therefore quite easy to distinguish (making multiple matches unlikely). The next experiment will have stochastic profiles.

The predictive features provided by the prediction profile models are clearly useful for control, as the control performance of OLGARB using their predictions approaches, and eventually exactly matches that of OLGARB using the true predictions (marked "True"). The inaccurate predictions provided by the POMDP were not very useful for control; OL-GARB using the POMDP provided predictions does not even break even, meaning it loses the game more often than it wins. The POMDP features did, however, seem to contain some useful information beyond that provided by the second-order Markov features (marked "SOM") which, as one might expect, performed very poorly.

## 6.4 Shooting Gallery

The second example is called the Shooting Gallery, pictured in Figure 7(a). The agent has a gun aimed at a fixed position on an $8 \times 8$ grid (marked by the $X$) . A target moves diagonally, bouncing off of the boundaries of the image and $2 \times 2$ obstacles (an example trajectory is pictured). The agent's task is to shoot the target. The agent has two actions: *watch* and *shoot*. When the agent chooses *watch*, it gets 0 reward. If the agent chooses *shoot* and the target is in the crosshairs in the step *after* the agent shoots, the agent gets reward of 10, otherwise it gets a reward of -5. Whenever the agent hits the target, the shooting range resets: the agent receives a special "reset" observation, each $2 \times 2$ square on the range is made an obstacle with probability 0.1, and the target is placed in a random position. There is also a 0.01 probability that the range will reset at every time step. The difficulty is that the target is "sticky." Every time step with probability 0.7 it moves in its current direction, but with probability 0.3 it sticks in place. Thus, looking only at recent history, the agent may not be able to determine the target's current direction. The agent needs to know the probability that the target will be in its sights in the next step, so clearly the single test of interest is: *watch target* (that is "If I choose the *watch* action, will the *target* enter the crosshairs?"). When the target is far from the crosshairs, the prediction of this test will be 0. When it target is in the crosshairs, it will be 0.3. When the target is

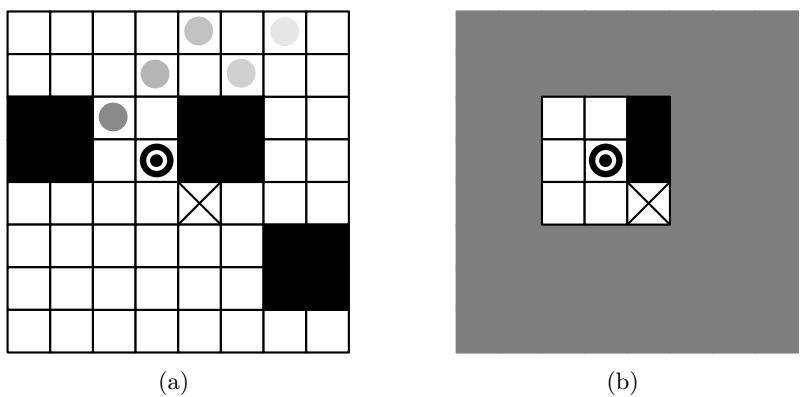

(a)                              (b)

Figure 7: The Shooting Gallery domain. (a) A possible arrangement of obstacles and tra-
jectory for the target (lighter is further back in time). In this case the target will
definitely not enter the agent's crosshairs, since it will bounce off of the obstacle.
(b) The abstraction applied to the most recent observation.

near the crosshairs, the model must determine whether the prediction is 0.7 or 0, based on
the target's previous behavior (its direction) and the configuration of nearby obstacles.

This problem has stochastic prediction profiles, so it is expected that more data will
be required to differentiate them. Also, due to the number of possible configurations of
obstacles and positions of the target, this system has roughly 4,000,000 observations and
even more latent states. This results in a large number of possible histories, each with only
a small probability of occurring. As discussed in Section 4, this can lead to a large sample
complexity for obtaining good estimates of prediction profiles. Here this is addressed with
a simple form of generalization: observation abstraction. Two observations are treated as
the same if the target is in the same position and if the configuration of obstacles in the
immediate vicinity of the target is the same. In other words, each abstract observation
contains information only about the target's position and the obstacles surrounding the
target, and not the placement of obstacles far away from the target (see Figure 7(b)) for an
example. Under this abstraction, the abstract observations still provide enough detail to
make accurate predictions. That is, two histories do indeed have the same prediction profile
if they have the same action sequence and their observation sequences correspond to the
same sequence of aggregate observations. This enables one sample trajectory to improve
the estimates for several histories, though, even with this abstraction, there are still over
2000 action-observation pairs. The same observation abstraction was applied when training
the POMDP model.

Training trajectories were length 4 and the search for profiles was restricted to length
3 histories. Results are shown in Figure 8. Perhaps the most eye-catching feature of
the results is the *upward* trending curve in the prediction error graph, corresponding to
the PP-POMDP with the KL-Divergence based matching (labeled "PP-POMDP(KLD)").
Recall that the danger of the KL-divergence based matching strategy is that it may produce
incorrect labels in the training data. Apparently these errors were severe enough in this
problem to drastically mislead the POMDP model. With a small amount of data it obtained very good prediction error, but with more data came more misleading labelings, and the performance suffered. The PP-POMDP trained with the other matching method ("PP-POMDP(cut)") displays a more typical learning curve (more data results in better error), though it takes a great deal of data before it begins to make reasonable predictions. This is because cutting off trajectories that have multiple matches throws away data that might have been informative to the model. The PP-LPSTs generally outperform the PP-POMDPs in this problem. With the trajectory cutting method, the PP-LPST ("PP-LPST(cut)") quickly outperforms the flat POMDP and, with enough data, outperforms both versions of PP-POMDP. The PP-LPST with the KL-divergence based matching ("PP-LPST(KLD)") is by far the best performer, quickly achieving small prediction error. Clearly the incorrect labels in the training data did not have as dramatic an effect on the LPST learning, possibly because, as a suffix tree, an LPST mostly makes its predictions based on recent history, limiting the effects of labeling errors to a few time-steps.

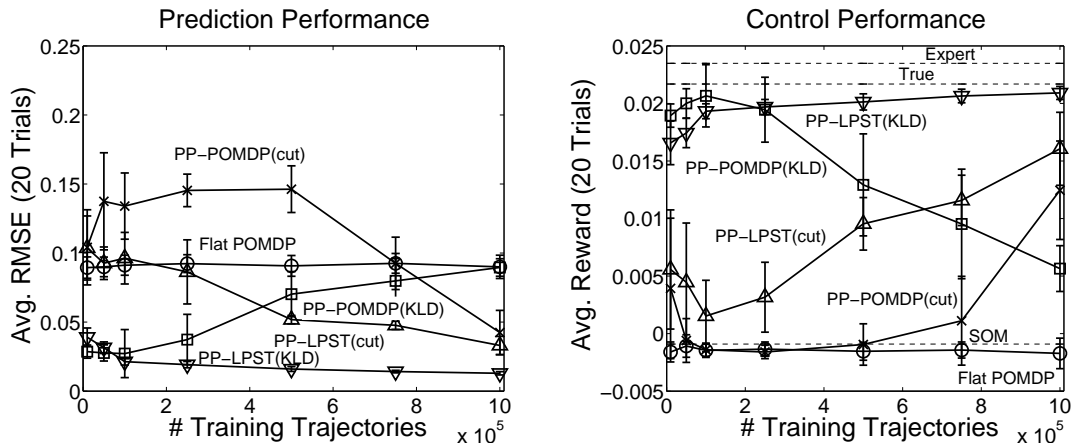

Figure 8: Results in the Shooting Gallery domain.

Control performance essentially mirrors prediction performance, with some interesting exceptions. Note that even though PP-POMDP(KLD) obtains roughly the same prediction error as the flat POMDP at 1,000,000 training trajectories, the predictive features it provides still result in substantially better control performance. This indicates that, even though the PP-POMDP is making errors in the exact values of the predictions, it has still captured more of the important dynamics of the predictions than the flat POMDP has. The flat POMDP itself provides features that are roughly as useful as second-order Markov features, which do not result in good performance. Again, OLGARB using these features does not break even, meaning it is wasting bullets when the target is not likely to enter the crosshairs. The best-performing prediction profile model, PP-LPST(KLD) approaches the performance of OLGARB using the true predictions with sufficient data.

## 7. Related Work

The idea of modeling only some aspects of the observations of a dynamical system has certainly been raised before. For instance, in a recent example Rudary (2008) learned linear-Gaussian models of continuous partially observable environments where some dimensions of the observation were treated as unmodeled "exogenous input." These inputs were assumed to have a linear effect on state transition. Along somewhat similar lines, but in the context of *model minimization* (taking a given, complete model and deriving a simpler, abstract model that preserves the value function) Wolfe (2010) constructed both an abstract model and a "shadow model" that predicts observation details that are ignored by the abstraction. The "shadow model" takes the abstract observations of the abstract model as unmodeled input. Splitting the observation into modeled and un-modeled components and then learning a generative model is certainly related to our approach. In that case, a model would make *all* conditional predictions about the modeled portion of the observation, given the exogenous inputs (as well as actual actions and the history). Prediction profile models take this to an extreme, by treating the *entire* observation as input. Instead of predicting future sequences of some piece of the next observation conditioned on another piece, prediction profile models predict the values of an arbitrary set of predictions of interest at the next time step, given the entire action and observation. This allows significantly more freedom in choosing which predictions the model will make (and, more importantly, will not make).

One modeling method closely related to prediction profiles is Causal State Splitting Reconstruction (CSSR) (Shalizi & Klinker, 2004). CSSR is an algorithm for learning generative models of discrete, partially observable, uncontrolled dynamical systems. The basic idea is to define an equivalence relation over histories where two histories are considered equivalent if they are associated with identical distributions over possible futures. The equivalence classes under this relation are called *causal states*. The CSSR algorithm learns the number of causal states, the distribution over next observations associated with each causal state, and the transitions from one causal state to the next, given an observation. It is straightforward to see that there is a one-to-one correspondance between causal states and the predictive states of a PSR. As such, a causal state model is precisely the prediction profile model where the set of tests of interest is $Q$, some set of core tests. With this correspondance in hand, the results in Section 5.2 show that in many cases the number of causal states will *greatly* exceed the linear dimension of the original system and that therefore CSSR may be inadvisable in many problems, in comparison to more standard modeling approaches. It is possible that the CSSR algorithm could be adapted to the more general setting of arbitrary sets of tests of interest, however the algorithm does rely heavily on the fact that a prediction profile model with $Q$ as the tests of interest is Markov, which is not generally the case for other sets of tests of interest.

As mentioned in Section 2, McCallum (1995) presented UTree, a suffix-tree-based algorithm for learning value functions in partially observable environments. Because UTree learns only the value function (a prediction about future rewards), and does not make any predictions about observations, UTree does learn a non-generative partial model. Wolfe and Barto (2006) extend UTree to make one-step predictions about particular observation features rather than limiting predictions to the value function. Because it learns a suffix tree, UTree is able to operate on non-episodic domains (whereas our method requires seeing histories multiple times) and is not required to explicitly search for distinct prediction profiles. UTree also directly incorporates abstraction learning, learning simultaneously which observation features are important, and where in the history suffix to attend to them. That said, the main drawback of the suffix tree approach is that the tree only takes into account information from relatively recent history (a suffix of the history). It cannot "remember" important information for an arbitrary number of steps as a recurrent state-based model can. In the Three Card Monte example, for instance, having access to a depth-limited suffix of history would be of little help. In order to track the ace, one must take into account every move the dealer has made since the beginning of the game. UTree would essentially forget where the card was if the game's length surpassed the depth of its memory.

McCallum (1993) and Mahmud (2010) both provide methods for learning state machines that predict the immediate reward resulting from any given action-observation pair in partially observable control tasks (and thus do not suffer from the issue of finite-depth memory that suffix trees do). Thus, their learning problem is a special case of ours, where they restrict their models to make one-step predictions about the immediate reward. In both cases, a simple model is incrementally and greedily elaborated by proposing states to be split and evaluating the results (via statistical tests in the case of McCallum and via likelihood hill-climbing in the case of Mahmud). McCallum expressed concern that his approach had difficulty extracting long-range dependencies (for instance, learning to attend to an event that does not appear to affect the distribution of rewards until many steps later); it is not clear the extent to which Mahmud's approach addresses this issue. These methods have some of the advantages of UTree, most notably that they can be applied to non-episodic domains. That said, our approach has advantages as well. By re-casting the problem of learning a non-generative model as a standard *generative* model-learning problem, we have been able to gain deeper understanding of the complexity and applicability of prediction profile models compared to more standard generative models. Furthermore, this has allowed us to incorporate standard, well-studied generative model-learning methods into our learning algorithm, thereby leveraging their strengths in the non-generative setting. Most specifically, this has resulting in a principled (albeit heuristic) learning algorithm, that does not rely on guess-and-check or stochastic local search.

The prediction profile system is also similar in spirit to finite state controllers for POMDPs. Sondik (1978) noted that in some cases, it is possible to represent the optimal policy for a POMDP as a finite state machine. These finite state controllers are very much like prediction profile models in that they take action-observation pairs as inputs, but instead of outputting predictions associated with the current history, they output the optimal action to take. Multiple authors (e.g., Hansen, 1998; Poupart & Boutilier, 2003) provide techniques for learning finite state controllers. However, these algorithms typically require access to a complete POMDP model of the world to begin with which, in our setting, is assumed to be impractical.

## 8. Conclusions and Future Directions

The most standard methods for learning models in partially observable environments learn generative models. If one has only a small set of predictions of interest to make (and therefore does not require the full power of a generative model), one can ignore irrelevant detail via abstraction to simplify the learning problem. Even so, a generative model will necessarily make predictions about any *relevant* details, even if they are not directly of interest. We have seen by example that the resulting model can be counter-intuitively complex, even if the predictions the model is being asked to make are quite simple.

We presented prediction profile models, which are *non-generative* models for partially observable systems that make *only* the predictions of interest and no others. The main idea of prediction profile models is to learn a model of the dynamics of the *predictions themselves* as they change over time, rather than a model of the dynamics of the system. The learning method for prediction profile models learns a transformation of the training data and then applies standard methods to the transformed data (assuming that the predictions of interest take on only a finite number of distinct values). As a result, it retains advantages of methods like EM for POMDPs that learn what information from history must be maintained in order to make predictions (rather than requiring a set of history features *a priori*). We showed that a prediction profile model can be far simpler than a generative model, though it can also be far more complex, depending on what predictions it is asked to make. However, if the predictions of interest depend on relatively little state information, prediction profile models can provide substantial savings over standard modeling methods such as POMDPs.

While the experiments in Section 6 demonstrate that it is possible to learn prediction profile models in contrived systems too complex for POMDPs, the specific learning algorithm presented here is not likely to scale to more natural domains without modification. The most critical scaling issues for prediction profile models are the sample complexity of estimating the prediction profiles, and the computational complexity of searching for prediction profiles and translating the data. In both cases, the critical source of complexity is essentially how many distinct histories there are in the training data (more distinct histories means the data is spread thin amongst them and there are more estimated profiles to search through). As such, generalization of prediction estimates across many histories would be a key step toward applying these ideas to more realistic domains. We are currently developing learning algorithms that combine the ideas behind prediction profile models with methods for learning abstractions that allow many essentially equivalent histories to be lumped together for the purposes of estimating the predictions of interest.

Another limitation of the prediction profile model learning method presented here is its reliance on the assumption of a finite number of prediction profiles. While this assumption does hold in many cases, an ideal method would be able to deal gracefully with a very large or infinite number of prediction profiles. One possibility is to simply cluster the predictions in other ways. For instance, one may only desire a certain level of prediction accuracy and may therefore be willing to lump some distinct prediction profiles together in exchange for a simpler prediction profile system. Another idea would be to learn a prediction profile model using continuous-valued representations such as Kalman filters (Kalman, 1960) or PLGs (Rudary, Singh, & Wingate, 2005) (or their nonlinear variants, e.g., Julier & Uhlmann, 1997; Wingate, 2008). These representations and learning algorithms explicitly deal with systems with an infinite number of observations (prediction profiles in this case). Even when there are finitely many prediction profiles, methods for learning non-linear continuous models may still be able to (approximately) capture the discrete dynamics.

Additionally, though our results have focused on discrete systems, the main motivation behind prediction profile models also has purchase in the continuous setting. Typical methods for learning models of partially observable systems in continuous systems, much like their discrete valued counterparts, learn *generative* models. As such, the non-generative approach of prediction profile models may provide similar benefits in the continuous setting if not all predictions need be made. In this setting, prediction profiles might be represented in a parametric form (for instance, the mean and variance of a Gaussian). The main idea of prediction profile models (though not the specific method presented here) could still then be applied: learn a model of the dynamics of these distribution parameters, rather than the dynamics of the system itself.

Finally, we have not discussed in this work *how* the tests of interest should be determined, only how to predict them once they are selected. Automatically selecting interesting/important predictive features as targets for partial models would certainly be an interesting research challenge. Of course, this would depend on what the predictions will be used for. If the predictions will be used as features for control, as we have done in our experiments, then it would certainly seem intuitive to start with predictive features regarding the reward signal, and perhaps observation features that strongly correlate with reward (as we have intuitively done by hand in our experiments). It may also be useful to consider making predictions *about those predictions* in the style of TD Networks (Sutton & Tanner, 2005). For instance, one could imagine learning models that make predictions about which profile another model will emit. In this way models could be chained together to make predictions about more extant rewards, rather than focusing solely on predicting the immediate reward signal (which is not always a particularly good feature for temporal decision problems). Another common use of partial models is to decompose a large modeling problem into many small ones, as in, for instance, factored MDPs (Boutilier et al., 1999), factored PSRs (Wolfe et al., 2008), or collections of local models (Talvitie & Singh, 2009b). In this setting, choosing tests of interest would be an example of the structure learning problem: decomposing one-step predictions into relatively independent components and then assigning them to different models.

## Acknowledgments

Erik Talvitie was supported under the NSF GRFP. Satinder Singh was supported by NSF grant IIS-0905146. Any opinions, findings, and conclusions or recommendations expressed in this material are those of the authors and do not necessarily reflect the views of the NSF.

The work presented in this paper is an extension of work presented at IJCAI (Talvitie & Singh, 2009a). We are grateful to the anonymous reviewers whose helpful comments have improved the presentation of this work.

## Appendix A.

### A.1 Proof of Proposition 15

This result will follow straightforwardly from a general fact about dynamical systems. Let $h^{[i...j]}$ be the sequence of actions and observations from $h$ starting with the $i$th time-step in the sequence and ending with the $j$th time-step in the sequence. For convenience's sake, if $i > j$ let $h^{[i...j]} = h^0$, the null sequence. The following two results will show that if some test $t$ *ever* has positive probability, then it must have positive probability at some history with length less than the linear dimension of the system.

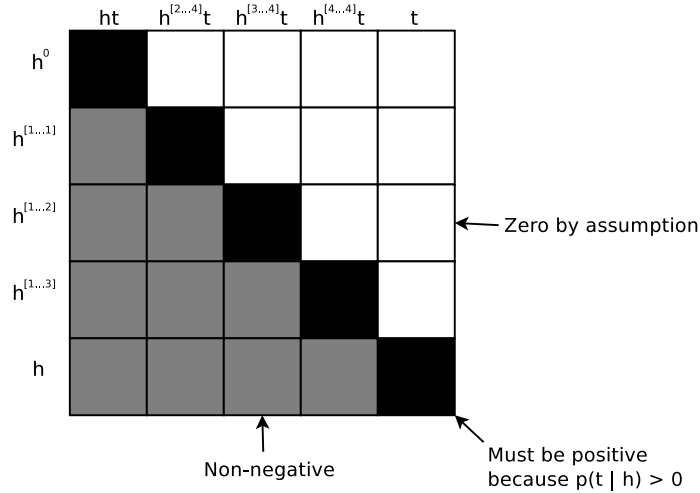

Figure 9: The matrix constructed in Lemma 21 is full rank (a contradiction).

**Lemma 21.** *If the linear dimension of a dynamical system is $n$, then for any test $t$ and history $h$ with $length(h) = k \geq n$ and $p(t \mid h) > 0$, $\exists i, j$ with $0 \leq i < j - 1 \leq k$ such that $p(t \mid h^{[1...i]}h^{[j...k]}) > 0$.*

*Proof.* Note that because $p(t \mid h) > 0$, $p(h^{[(i+1)...k]}t \mid h^{[1...i]}) = p(t \mid h)p(h^{[(i+1)...k]} \mid h^{[1...i]}) > 0$ for all $0 \leq i \leq k$. Now assume for all $i, j$ with $0 \leq i < j - 1 \leq k$ that $p(h^{[j...k]}t \mid h^{[1...i]}) = p(t \mid h^{[1...i]}h^{[j...k]})p(h^{[j...k]} \mid h^{[1...i]}) = 0$ and seek a contradiction. Consider a submatrix of the system dynamics matrix. The rows of this submatrix correspond to prefixes of $h$: $h^{[1...i]}$ for all $0 \leq i \leq k$. The columns correspond to suffixes of $h$ pre-pended to the test $t$: $h^{[j...k]}t$ for all $1 \leq j \leq k + 1$. This is a $k + 1 \times k + 1$ matrix. Under the above assumption, this matrix is triangular with positive entries along the diagonal (Figure 9 shows this matrix when $k = 4$). As such, this matrix is full rank (rank $k + 1$). This is a contradiction since $k \geq n$ and a submatrix can never have higher rank than the matrix that contains it. ☐

The next result follows immediately from Lemma 21.

**Corollary 22.** *If the system has linear dimension $n$ and for some test $t$ and history $h$ $p(t \mid h) > 0$, then there exists a (possibly non-consecutive) subsequence $h'$ of $h$ such that $length(h') < n$ with $p(t \mid h') > 0$.*

*Proof.* By Lemma 21, every history $h$ with length $k \geq n$ such that $p(t \mid h) > 0$ must have a subsequence $h_1$ with length $k_1 < k$ such that $p(t \mid h) > 0$. If $k_1 \geq n$, then $h_1$ must have a subsequence $h_2$ with length $k_2 < k_1$. This argument can be repeated until the subsequence has length less than $n$. ☐

The consequence of Corollary 22 is that *every* test that ever has positive probability, must have positive probability following some history of length less than $n$. With this fact in hand, Proposition 15 can now be proven.

**Proposition 15.** *For any deterministic dynamical system with actions $\mathcal{A}$, and observations $\mathcal{O}$, the linear dimension, $n \geq \frac{\log(|\mathcal{A}|-1)+\log(|\mathcal{O}|+1)}{\log|\mathcal{A}|}$.*

*Proof.* Since the system is deterministic, each history and action correspond to exactly one resulting observation. A history is a sequence of actions and observations. However, since the sequence of observations is fully determined by the sequence of actions in a deterministic system, the number of distinct histories of length $k$ is simply $|\mathcal{A}|^k$. At each history there are $|\mathcal{A}|$ action choices that could each result in a different observation. So, the number of observations that could possibly occur after histories of length $k$ is simply $|\mathcal{A}|^{k+1}$. By Corollary 22, if the linear dimension is $n$, *all* observations must occur after some history $h$ with $length(h) \leq n-1$. Thus, the number of observations that can possibly follow histories of length less than $n$ is:

$$|\mathcal{O}| \leq \sum_{i=0}^{n-1} |\mathcal{A}|^{i+1} = \frac{|\mathcal{A}|^{n+1} - 1}{|\mathcal{A}| - 1} - 1.$$

Solving for $n$ yields the bound on linear dimension in terms of the number of actions and the number of observations. □

## A.2 Proof of Proposition 17

**Proposition 17.** *For a given system and set of tests of interest, the linear dimension of the corresponding prediction profile system is no greater than that of the prediction profile system associated with any set of core tests for the system (as described in Section 2.2).*

*Proof.* Recall from the discussion of PSRs in Section 2.2 that a set of core tests, $Q$, is a set of tests whose corresponding columns in the system dynamics matrix constitute a basis. The predictions for the core tests at a given history form the *predictive state* at that history. So, the predictive state is precisely the prediction profile for the core tests $Q$. The prediction for *any* other test can be computed as a linear function of the prediction profile for $Q$. Note that the prediction profile system for $Q$ is itself an MDP. It was shown in Section 2.2 how to compute the next predictive state given the current predictive state and an action-observation pair.

Now consider some other set of tests of interest $\mathcal{T}^I$. Because the predictions for $Q$ can be used to compute the prediction for any other test, it must be that there is some function $\zeta$ that maps the prediction profiles for $Q$ to the prediction profiles for $\mathcal{T}^I$. In general, multiple predictive states may map to the same prediction profile for $\mathcal{T}^I$ so $\zeta$ is a surjection. Now it is easy to see that the prediction profile system for $\mathcal{T}^I$ is the result of applying the observation abstraction $\zeta$ to the prediction profile system for $Q$. Performing observation abstraction on an MDP generally produces a POMDP, but *never* increases the linear dimension (Talvitie, 2010). Hence, the prediction profile system for any set of tests of interest $\mathcal{T}^I$ has linear dimension no greater than that of the prediction profile system for any set of core tests, $Q$. □

## A.3 Proof of Proposition 20

**Proposition 20.** *Consider a POMDP with hidden states $\mathcal{S}$, actions $\mathcal{A}$, and observations $\mathcal{O}$. Let $\mathcal{T}^I$ be the set of tests of interest. Let $a^i$ be the action taken at time-step $i$, $s^i$ be the hidden state reached after taking action $a^i$, and $o^i$ be the observation emitted by $s^i$. Now,* *consider* any *surjection* $\sigma : \mathcal{S} \to \mathcal{S}^{\sigma}$ *mapping hidden states to a set of* abstract states *with the following properties:*

1. *For any pair of primitive states $s_1, s_2 \in \mathcal{S}$, if $\sigma(s_1) = \sigma(s_2)$, then for any time-step $i$ and any test of interest $t \in \mathcal{T}^I$, $p(t \mid s^i = s_1) = p(t \mid s^i = s_2)$.*

2. *For any pair of primitive states $s_1, s_2 \in \mathcal{S}$, if $\sigma(s_1) = \sigma(s_2)$, then for any time-step $i$, abstract state $S \in \mathcal{S}^{\sigma}$, observation $o \in \mathcal{O}$, and action $a \in \mathcal{A}$,*

$$Pr(\sigma(s^{i+1}) = S \mid s^i = s_1, a^{i+1} = a, o^{i+1} = o) =$$
$$Pr(\sigma(s^{i+1}) = S \mid s^i = s_2, a^{i+1} = a, o^{i+1} = o).$$

*If such a $\sigma$ exists, then the prediction profile system for $\mathcal{T}^I$ has linear dimension no greater than the number of distinct beliefs over abstract states, $\mathcal{S}^{\sigma}$.*

*Proof.* The proof follows similar reasoning to the proof of Proposition 17. Note that, because of Property 1 the belief over abstract states at a given history is sufficient to compute the prediction profile. For any history $h$ and any test of interest $t \in \mathcal{T}^I$:

$$p(t \mid h) = \sum_{s \in \mathcal{S}} \Pr(s \mid h) p(t \mid s) = \sum_{S \in \mathcal{S}^{\sigma}} \sum_{s \in S} \Pr(s \mid h) p(t \mid s)$$
$$= \sum_{S \in \mathcal{S}^{\sigma}} p(t \mid S) \sum_{s \in S} \Pr(s \mid h) = \sum_{S \in \mathcal{S}^{\sigma}} p(t \mid S) \Pr(S \mid h),$$

where the third equality follows from property 1: for any $S \in \mathcal{S}^{\sigma}$, all hidden states $s \in S$ have the same associated probabilities for the tests of interest.

Now, consider the dynamical system with beliefs over abstract states as "observations" and action-observation pairs as "actions." Call this the *abstract belief system.* Just as with the predictive state, because it is possible to compute the prediction profile from the abstract beliefs, the prediction profile model for $\mathcal{T}^I$ can be seen as the result of an observation aggregation of the abstract belief system. As a result, the prediction profile system has linear dimension no greater than that of the abstract belief system.

The rest of the proof shows that, because of Property 2, the abstract belief system is an MDP, and therefore has linear dimension no greater than the number of distinct beliefs over abstract states.

Given the probability distribution over abstract states at a given history $h$, and the agent takes an action $a$ and observes and observation $o$, it is possible to compute the probability of an abstract state $S \in \mathcal{S}^{\sigma}$ at the new history:

$$\Pr(S \mid hao) = \sum_{s \in \mathcal{S}} \Pr(s \mid h) \Pr(S \mid s, a, o) = \sum_{S' \in \mathcal{S}^{\sigma}} \sum_{s \in S'} \Pr(s \mid h) \Pr(S \mid s, a, o)$$
$$= \sum_{S' \in \mathcal{S}^{\sigma}} \Pr(S \mid S', a, o) \sum_{s \in S'} \Pr(s \mid h) = \sum_{S' \in \mathcal{S}^{\sigma}} \Pr(S \mid S', a, o) \Pr(S' \mid h),$$

where the third equality follows from Property 2: for any $S \in \mathcal{S}^{\sigma}$, all hidden states $s \in S$ have the same associated conditional distribution over next abstract states, given the action and observation.

So, because one can compute the next abstract beliefs from the previous abstract beliefs, the abstract belief system is an MDP, and therefore has linear dimension no greater than the number of observations (the number of distinct abstract beliefs). Because one can compute the prediction profile from the abstract beliefs, the prediction profile system can be constructed by applying an observation abstraction to the abstract belief system. Thus, the prediction profile system has linear dimension no greater than the number of distinct abstract beliefs. □

# References

Baum, L. E., Petrie, T., Soules, G., & Weiss, N. (1970). A maximization technique occuring in the statistical analysis of probabilistic functions of markov chains. *The Annals of Mathematical Statistics*, *41*(1), 164–171.

Baxter, J., & Bartlett, P. L. (2000). Reinforcement learning in POMDPs via direct gradient ascent. In *Proceedings of the Eighteenth International Conference on Machine Learning (ICML)*, pp. 41–48.

Boots, B., Siddiqi, S., & Gordon, G. (2010). Closing the learning-planning loop with predictive state representations. In *Proceedings of Robotics: Science and Systems*, Zaragoza, Spain.

Boots, B., Siddiqi, S., & Gordon, G. (2011). An online spectral learning algorithm for partially observable nonlinear dynamical systems. In *Proceedings of the Twenty-Fifth National Conference on Artificial Intelligence (AAAI)*.

Boutilier, C., Dean, T., & Hanks, S. (1999). Decision-theoretic planning: Structural assumptions and computational leverage. *Journal of Artificial Intelligence Research*, *11*, 1–94.

Bowling, M., McCracken, P., James, M., Neufeld, J., & Wilkinson, D. (2006). Learning predictive state representations using non-blind policies. In *Proceedings of the Twenty-Third International Conference on Machine Learning (ICML)*, pp. 129–136.

Dinculescu, M., & Precup, D. (2010). Approximate predictive representations of partially observable systems. In *Proceedings of the Twenty-Seventh International Conference on Machine Learning (ICML)*, pp. 895–902.

Hansen, E. (1998). *Finite-Memory Control of Partially Observable Systems*. Ph.D. thesis, University of Massachussetts, Amherst, MA.

Holmes, M., & Isbell, C. (2006). Looping suffix tree-based inference of partially observable hidden state. In *Proceedings of the Twenty-Third International Conference on Machine Learning (ICML)*, pp. 409–416.

James, M., & Singh, S. (2004). Learning and discovery of predictive state representations in dynamical systems with reset. In *Proceedings of the Twenty-First International Conference on Machine Learning (ICML)*, pp. 417–424.

Julier, S. J., & Uhlmann, J. K. (1997). A new extension of the kalman filter to nonlinear systems. In *Proceedings of AeroSense: The Eleventh International Symposium on Aerospace/Defense Sensing, Simulation and Controls*, pp. 182–193.

Kalman, R. E. (1960). A new approach to linear filtering and prediction problems. *Transactions of the ASME – Journal of Basic Engineering, 82*, 35–45.

Littman, M., Sutton, R., & Singh, S. (2002). Predictive representations of state. In *Advances in Neural Information Processing Systems 14 (NIPS)*, pp. 1555–1561.

Littman, M. L. (1996). *Algorithms for Sequential Decision Making*. Ph.D. thesis, Brown University, Providence, RI.

Mahmud, M. M. H. (2010). Constructing states for reinforcement learning. In *Proceedings of the Twenty-Seventh International Conference on Machine Learning (ICML)*, pp. 727–734.

McCallum, A. K. (1995). *Reinforcement Learning with Selective Perception and Hidden State*. Ph.D. thesis, Rutgers University.

McCallum, R. A. (1993). Overcoming incomplete perception with utile distinction memory. In *Proceedings of the Tenth International Conference on Machine Learning (ICML)*, pp. 190–196.

Monahan, G. E. (1982). A survey of partially observable markov decisions processes: Theory, models, and algorithms. *Management Science, 28*(1), 1–16.

Peters, J., & Schaal, S. (2008). Natural actor-critic. *Neurocomputing, 71*, 1180–1190.

Poupart, P., & Boutilier, C. (2003). Bounded finite state controllers. In *Advances in Neural Information Processing Systems 16 (NIPS)*.

Puterman, M. L. (1994). *Markov Decision Processes: Discrete Stochastic Dynamic Programming*. John Wiley and Sons, New York, NY.

Rivest, R. L., & Schapire, R. E. (1994). Diversity-based inference of finite automata. *Journal of the Association for Computing Machinery, 41*(3), 555–589.

Rudary, M. (2008). *On Predictive Linear Gaussian Models*. Ph.D. thesis, University of Michigan.

Rudary, M., Singh, S., & Wingate, D. (2005). Predictive linear-gaussian models of stochastic dynamical systems. In *Uncertainty in Artificial Intelligence: Proceedings of the Twenty-First Conference (UAI)*, pp. 501–508.

Shalizi, C. R., & Klinker, K. L. (2004). Blind construction of optimal nonlinear recursive predictors for discrete sequences. In *Proceedings of the Twentieth Conference on Uncertainty in Artificial Intelligence (UAI)*, pp. 504–511.

Singh, S., James, M. R., & Rudary, M. R. (2004). Predictive state representations: A new theory for modeling dynamical systems. In *Uncertainty in Artificial Intelligence: Proceedings of the Twentieth Conference (UAI)*, pp. 512–519.

Sondik, E. J. (1978). The optimal control of partially observable markov processes over the infinite horizon: Discounted costs. *Operations Research, 26*, 282–304.

Soni, V., & Singh, S. (2007). Abstraction in predictive state representations. In *Proceedings of the Twenty-Second National Conference on Artificial Intelligence (AAAI)*, pp. 639–644.

Sutton, R. S., & Tanner, B. (2005). Temporal-difference networks. In *Advances in Neural Information Processing Systems 17 (NIPS)*, pp. 1377–1384.

Talvitie, E. (2010). *Simple Partial Models for Complex Dynamical Systems*. Ph.D. thesis, University of Michigan, Ann Arbor, MI.

Talvitie, E., & Singh, S. (2009a). Maintaining predictions over time without a model. In *Proceedings of the Twenty-First International Joint Conference on Artificial Intelligence (IJCAI)*, pp. 1249–1254.

Talvitie, E., & Singh, S. (2009b). Simple local models for complex dynamical systems. In *Advances in Neural Information Processing Systems 21 (NIPS)*, pp. 1617–1624.

Weaver, L., & Tao, N. (2001). The optimal reward baseline for gradient-based reinforcement learning. In *Uncertainty in Artificial Intelligence: Proceedings of the Seventeenth Conference (UAI)*, pp. 538–545.

Williams, R. (1992). Simple statistical gradient-following algorithms for connectionist reinforcement learning. *Machine Learning, 8*, 229–256.

Wingate, D. (2008). *Exponential Family Predictive Representations of State*. Ph.D. thesis, University of Michigan.

Wingate, D., Soni, V., Wolfe, B., & Singh, S. (2007). Relational knowledge with predictive state representations. In *Proceedings of the Twentieth International Joint Conference on Artificial Intelligence (IJCAI)*, pp. 2035–2040.

Wolfe, A. P. (2010). *Paying Attention to What Matters: Observation Abstraction in Partially Observable Environments*. Ph.D. thesis, University of Massachussetts, Amherst, MA.

Wolfe, A. P., & Barto, A. G. (2006). Decision tree methods for finding reusable MDP homomorphisms. In *Proceedings of the Twenty-First National Conference on Artificial Intelligence (AAAI)*.

Wolfe, B., James, M., & Singh, S. (2008). Approximate predictive state representations. In *Proceedings of the Seventh Conference on Autonomous Agents and Multiagent Systems (AAMAS)*.

Wolfe, B., James, M. R., & Singh, S. (2005). Learning predictive state representations in dynamical systems without reset. In *Proceedings of the Twenty-Second International Conference on Machine Learning (ICML)*, pp. 985–992.

Wolfe, B., & Singh, S. (2006). Predictive state representations with options. In *Proceedings of the Twenty-Third International Conference on Machine Learning (ICML)*, pp. 1025–1032.